%% file: conference_101719.tex
\documentclass[compsoc, conference, a4paper, 10pt, times]{IEEEtran}
\IEEEoverridecommandlockouts
\usepackage{cite}
\usepackage{amsmath,amssymb,amsfonts}
\usepackage{algorithmic}
\usepackage{graphicx}
\usepackage{textcomp}
\usepackage{dirtytalk}
\usepackage{subcaption}
\usepackage{booktabs}
\usepackage{hyperref}
\usepackage{relsize}
\usepackage[table,svgnames,dvipsnames]{xcolor}
\usepackage[column=0]{cellspace}
\usepackage{tcolorbox}
\usepackage{makecell}
\usepackage{subcaption}
\usepackage{dsfont}
\usepackage[capitalize]{cleveref}
\usepackage{pgfplots}
\usepgfplotslibrary{statistics}
\usepackage{tikz}
\usetikzlibrary{arrows}

\newcommand{\E}[0]{\mathlarger{\mathbb{E}}}

\definecolor{myred}{HTML}{80131C}
\definecolor{myrose}{HTML}{D9969B}
\definecolor{mydarkblue}{HTML}{003554}
\definecolor{mylightblue}{HTML}{63A6C6}

\def\BibTeX{{\rm B\kern-.05em{\sc i\kern-.025em b}\kern-.08em
    T\kern-.1667em\lower.7ex\hbox{E}\kern-.125emX}}
\begin{document}

\title{SoK: Memorisation in Machine Learning
}



\author{\IEEEauthorblockN{Dmitrii Usynin \textsuperscript{1,2}, Moritz Knolle \textsuperscript{2,3}, Georgios Kaissis \textsuperscript{1,2,3}}
\IEEEauthorblockA{\textit{\textsuperscript{1}Department of Computing}
\textit{Imperial College London}
}
\IEEEauthorblockA{\textit{\textsuperscript{2}Institute for AI in Medicine} 
\textit{Technical University of Munich}
}
\IEEEauthorblockA{\textit{\textsuperscript{3}Konrad Zuse School for Excellence in Reliable AI} 
}
}

\maketitle
\thispagestyle{plain}
\pagestyle{plain}

\begin{abstract}
Quantifying the impact of individual data samples on machine learning models is an open research problem.
This is particularly relevant when complex and high-dimensional relationships have to be learned from a limited sample of the data generating distribution, such as in deep learning.
It was previously shown that, in these cases, models rely not only on extracting patterns which are helpful for generalisation, but also seem to be required to incorporate some of the training data more or less \textit{as is}, in a process often termed \textit{memorisation}.
This raises the question: if some memorisation is a requirement for effective learning, what are its privacy implications?
In this work we unify a broad range of previous definitions and perspectives on memorisation in ML, discuss their interplay with model generalisation and their implications of these phenomena on data privacy.
Moreover, we systematise methods allowing practitioners to detect the occurrence of memorisation or quantify it and contextualise our findings in a broad range of ML learning settings.
Finally, we discuss memorisation in the context of privacy attacks, differential privacy (DP) and adversarial actors.
\end{abstract}


\section{Introduction}
Machine learning (ML) models require access to large amounts of high-quality, diverse and well-curated data to perform well in a variety of tasks ranging from biomedical imaging \cite{seo2020machine, rueckert2019model} to text generation using large language models (LLMs) \cite{nijkamp2022codegen, taylor2022galactica, alayrac2022flamingo}.
Such data is often sensitive in nature, mandating that its disclosure be avoided.
However, with the advent of generative ML, it has become apparent that models often reproduce samples from their training datasets almost verbatim \cite{ippolitopreventing, carlini2019secret, carlini2021extracting}, thus posing potential privacy risks for data owners.
This phenomenon is not new to the ML privacy community, as it has been known that models trained without the use of privacy-enhancing techniques such as differential privacy (DP) are prone to inferences about their training data, such as membership inference (MIA) \cite{shokri2017membership} or data reconstruction attacks \cite{zhu2019deep}.
However, these observations in generative models have led to a recent spike in research trying to uncover (1) whether and to what extent models actually \textit{contain} their training data (to then regurgitate it) and (2) which training samples are more prone to this.

The aforementioned phenomenon is (often informally) termed \textit{training data memorisation}.
Memorisation is often viewed (and taught) as being the opposite pole of generalisation (that is, committing data samples to storage rather than learning general patterns from the training data).  
Obviously, this distinction is somewhat artificial, as it is all but impossible to delineate where general patterns end and the storage of actual data samples begins.
Beyond this fact, it has empirically been shown that --while models are able to learn data representations which are useful to the task at hand-- they are also entirely capable of fitting random input-output associations such as purely random labels in the context of supervised learning \cite{zhang2017understanding}.
In addition, while supervised ML models are usually able to perfectly fit to their training data, this often does not translate to a commensurate generalisation performance on unseen data.

Additionally, certain types of ML models, such as support vector machines or $k$-nearest neighbours, exhibit a learning process which is in its entirety based on storing data samples followed by a subsequent look-up process.
It thus becomes evident that the delineation between memorisation and generalisation may --in fact-- be moot. 
To the contrary, it appears that \emph{both}, memorisation and generalisation, are crucial to the learning process of ML models.
Prior works lend credence to this fact. 
For example, \cite{brown2021memorization} find that specific tasks can only be successfully learned through memorisation of portions of the training data. 
Moreover, the seminal works of \cite{feldman2020neural, feldman2020does} demonstrate that memorisation and generalisation not only co-occur, but that the former is actually a prerequisite for the latter if one aims to train a model of (close-to) optimal utility.
In addition, several prior works equate memorisation to overfitting, the phenomenon where, during training, the model's performance on unseen data starts deteriorating after initially having increased, pointing to an over-accommodation of the model to its training data.

These introductory remarks reveal that there is not only a lack of terminological clarity when discussing memorisation and generalisation in ML, but also that there exist a multitude of --partially conflicting-- definitions.
We thus identify a requirement for a systematisation of prior works describing these phenomena in various sub-domains of ML.
Moreover, we contend that such a systematisation must also tackle specific related points of importance:
\begin{itemize}
    \item Pinpointing which samples are (more) prone to memorisation;
    \item Quantifying, rather than merely detecting the presence of memorisation;
    \item Identifying the implications of memorisation in terms of data privacy;
    \item Discussing techniques aimed at preventing or diminishing memorisation and their consequences on the model's performance.
\end{itemize}

Such a work will not only aid in the disambiguation of the multiple existent definitions of memorisation, but also assist practitioners in choosing the techniques to identify, measure and possibly reduce memorisation in their operational scenario or use-case.

In this paper, we attempt the aforementioned systematisation.
Our work is structured as follows:

\begin{itemize}
    \item We outline a formalised definition of memorisation in machine learning and discuss how it can be quantified directly;
    \item In addition, we discuss notions which are ostensibly related to (but otherwise disjoint from) memorisation;
    \item We then discuss how memorisation can be actively induced, especially in generative ML models;
    \item Finally, we discuss the implications of memorisation on the privacy of the individuals whose data is used to train the model.
\end{itemize}

\section{Preliminaries}
Most ML models can be viewed as parameterised functions $f_\theta(\cdot)$, which, given some input $x$ and parameter values $\theta$, produce a corresponding output $f_\theta(x)$. 
Using a training dataset $S \sim \mathcal{D}$, where $\mathcal{D}$ is the data-generating distribution, many ML algorithms can be reduced to finding an optimal parameter setting $\theta^\ast$ (and the corresponding function $f_{\theta^\ast}$) that minimises a loss function $L$ over $S$:
\begin{equation}
    \label{eq:ERM}
    \theta^\ast := \underset{\theta}{\arg\min} \sum_{i=1}^{n} L(f_\theta(x_i), y_i),
\end{equation}
which is often called empirical risk minimisation (ERM).
Based on the form that $S$ takes, we now broadly categorize ML methods into: (I) \textit{supervised learning}, where the training dataset $S$ contains both inputs $x$ as well as ground-truth outputs/labels $y$ and (II) \textit{generative learning/modelling}, where no corresponding example outputs are available, but we wish to learn the general structure of $x$.
We use the terms \textit{supervised} and \textit{discriminative} interchangeably in this work.

For the rest of this paper, will omit $\theta$ (unless explicitly necessary) and denote the result of this procedure (commonly referred to as model training/fitting) as applying training algorithm $A$ to the dataset $S$ as $f \leftarrow A(S)$.

Finally, we frame the goal of ML as obtaining a model $f$ that generalises to unseen data.
This is measured by the generalisation performance/error, which for supervised learning is defined as follows:
\begin{equation}
    \label{eq:gen_error}
    \text{err}_{\text{gen}}(f) := \underset{(x,y) \sim \mathcal{D}}{\E}[L(f(x), y)].
\end{equation}
For the generative learning setting, $L$ takes only one input, as $S$ contains no labels $y$.
Note that formally, generalisation performance must be measured over the entirety of the data-generating distribution $\mathcal{D}$.
In practice, it can only be estimated as an empirical average over a (finite) test set.
Accurate estimation of $\text{err}_{\text{gen}}$ thus hinges on the choice of a representative test set that is distinct from the training dataset.
Throughout, we will use the terms \textit{sample}, \textit{instance} and \textit{data point} synonymously.

\section{A formal definition of memorisation} 
\label{sec:feldman}
For a long time, memorisation lacked a precise definition and the term was commonly used loosely to refer to a variety of phenomena. 
In this section, we discuss the definition of Feldman \cite{feldman2020does}, who presented the first unified formulation and theory of memorisation in ML.

In the framework of Feldman (developed further in \cite{feldman2020neural, zhang2021counterfactual}), memorisation is framed as the impact a particular sample has on its own prediction (known as the \textit{self-influence}).
Formally, it is defined as the difference in (expected) performance on the sample (with index) $i$ when sample $i$ is included in the training dataset $S$:
\begin{align}
    \mathrm{mem}&(A, S, i) := \notag \\
    & \underbrace{\underset{f \leftarrow A(S)}{\E}\left[\mathrm{M}\left( f(x_i),y_i \right)\right]}_{\color{myred}\textbf{performance on $i$ when $i \in S$}} - \underbrace{\underset{f \leftarrow A(S{\backslash i})}{\E}\left[\mathrm{M}\left(f(x_i),y_i\right) \right]}_{\color{myrose}\textbf{performance on $i$ when $i \notin S$}}.
    \label{eq:self-inf}
\end{align}
Here, the expectation is taken over the randomness of the training algorithm $A$. $\mathrm{M}$ refers to some suitable performance metric, (e.g. accuracy) and $S{\backslash i}$ is the training dataset with sample $i$ removed. Note that \cref{eq:self-inf} can be applied to generative settings by picking a performance metric $M$ that only takes a single input argument (i.e. without including the ground truth labels).
Note that a naive calculation of \cref{eq:self-inf} is computationally expensive. 
However, efficient sub-sampling estimators were proposed and are discussed in \cref{sec:influence}.

\begin{tcolorbox}[colback=gray!10!white, colframe=white!50!black, title=Key Point]
    \textit{Memorisation} is formally defined as the influence a sample has on its own (correct) prediction (self-influence).
\end{tcolorbox}

\subsection{Influence and the long-tail theory}

\begin{figure*}[tp]
    \begin{subfigure}[t]{0.6\linewidth}
        \centering
        \input{figs/long_tail}
        \caption{a long-tailed data distribution}
        \label{fig:Feldmans_long_tail}
    \end{subfigure}
    \begin{subfigure}[t]{0.36\linewidth}
        \centering
        \includegraphics[width=0.9\linewidth]{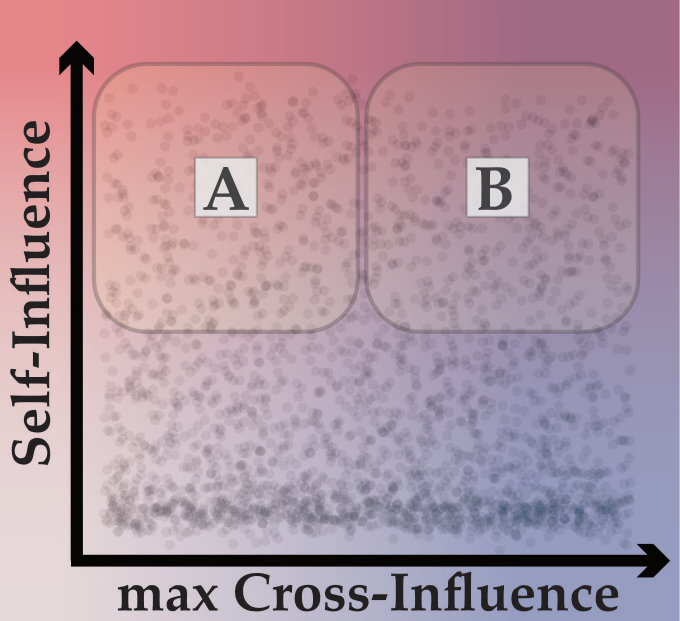}
        \caption{memorisation-influence continuum}
        \label{fig:influence_map}
    \end{subfigure}
    \caption{Long-tailed distributions have probability density that decays very slowly for extreme (atypical) values (a). 
    It can thus produce samples that are so distinct from typical samples that they are required to be memorised to allow for generalisation to other similar samples. 
    Sub-figure (b) shows a schematic visualisation of the memorisation-influence continuum. 
    While all samples from the tail of the data distribution are likely memorised during training (large self-influence), useless, low-quality or mislabeled samples (A) are indistinguishable from useful samples that significantly influence the correct prediction of at least one test sample (large max cross-influence) (B).
    }
\end{figure*}

In the same work \cite{feldman2020does}, Feldman theorised that memorisation is a required component of learning in specific learning settings.
This \textit{long-tail theory}, contends that close-to-optimal generalisation performance on long-tailed data distributions (see \cref{fig:Feldmans_long_tail}) necessitates the memorisation of rare and \say{atypical} samples from the tail of the distribution. 
This is due to the fact that, in long-tailed distributions, samples from low-density regions (the tail) differ extremely from samples from high-density regions (the head). 
Thus, memorisation of these atypical samples is necessary to generalise optimally to other atypical samples at test time.

The long-tail theory (developed further in \cite{feldman2020neural, brown2021memorization}), was supported by prior observations that naturally occurring data distributions commonly have long tails \cite{babbar2019data, zhu2014capturing, van2017devil, yang2022survey}.
Further empirical evidence in direct support of this theory was presented in \cite{feldman2020neural, zhang2021counterfactual}, which demonstrated that (for commonly used large-scale vision and text datasets), a substantial proportion of training examples have an out-sized impact on the model's performance on few specific test examples. 
The (sample-level) measure used to demonstrate this phenomenon was termed \textit{cross-influence}, that is, the impact a training sample $x_i$ has on the prediction of a test sample $x_j'$: 

\begin{align}
    \label{eq:cross-infl}
    \mathrm{infl}&(A, S, i, j) := \notag\\ 
    & \underbrace{\underset{{f \leftarrow A(S)}}{\E}[\mathrm{M}(f(x_{j}'),y_{j}')]}_{\color{mydarkblue}\textbf{performance on $j$ when $i \in S$}} - \underbrace{\underset{f \leftarrow A(S{\backslash i})}{\E}[\mathrm{M}(f(x_{j}'),y_{j}')]}_{\color{mylightblue}\textbf{performance on $j$ when $i \notin S$}},
\end{align}
whereby self-influence ($\mathrm{mem}$) is recovered when $i=j$ and $M$ is some suitable performance metric that returns higher values for better predictions. Thus, a positive $\mathrm{infl}$ value indicates that training sample $i$ improves the prediction on test sample $j$ when $i\in S$. If $M$ is taken to be a risk/loss function, equivalent behaviour can be achieved by simply negating all $\mathrm{infl}$ values.

\cref{fig:influence_map} visualises the memorisation-influence continuum.
The continuum illustrates that while all samples from the tail of the distribution are likely to be memorised (and thus exhibit high self-influence), not all of them have a significant impact on at least one test sample (A vs. B).
Further, a direct consequence of the long-tailed nature of data distributions is the fact that, at training time, low-quality or mislabelled samples (A) are statistically indistinguishable from useful representative samples of rare sub-populations (B) \cite{feldman2020does}. 
Note that, the data points in \cref{fig:influence_map} are drawn for illustrative purposes only. 

\begin{tcolorbox}[colback=gray!10!white, colframe=white!50!black, title=Key Point]
    When a data distribution is \textit{long-tailed}, memorising samples from the tail of the distribution can help a model generalise.
\end{tcolorbox}

\subsection{Prior efforts to capture memorisation}
\noindent
\textbf{Memorisation as overfitting}
\label{sec:overfitting}
A number of prior works \cite{olatunji2021membership,chen2020gan,kuppa2021towards,leino2020stolen,veale2018algorithms,hilprecht2019monte, mehta2020extreme} draw direct connections between memorisation and overfitting, defining memorisation of a given model as its generalisation gap (i.e. the empirical measure of overfitting).
Formally, the generalisation gap of a model $f$ is defined as:
\begin{align}    
    \text{err}_{\text{gap}}(f) := \underbrace{\text{err}_{\text{gen}}(f)}_{\text{generalisation error}} - \underbrace{\sum_{i=1}^n L(f(x_i), y_i)}_{\text{empirical (training) error on } S}
\end{align}
This quantity is task-agnostic and it can be estimated using a representative test set, making it a very attractive tool for empirical memorisation measurement.
However, in the light of evidence from \cite{feldman2020neural, zhang2021counterfactual}, showing that the memorisation of specific training examples actually increases generalisation performance, this definition now appears dated.
In addition, the measure above does not allow for the identification of individual memorised samples, being computed as an average.
Finally, recent works demonstrate that memorisation usually precedes overfitting \cite{carlini2019secret, tirumala2022memorization}.

\begin{tcolorbox}[colback=gray!10!white, colframe=white!50!black, title=Key Point]
Memorisation \textbf{cannot} be reduced to overfitting; in fact, memorisation and overfitting are distinct phenomena.  
\end{tcolorbox}
\noindent
\textbf{Fitting random labels}
Zhang et al. \cite{zhang2017understanding} demonstrated that even relatively small neural networks (by today's standards) can perfectly fit large datasets with randomised labels or even completely random data.
As the correct prediction of a random label is impossible without memorisation, randomised labels are commonly used to study qualitative aspects of memorisation behaviour in ML models, such as whether it can be localised to specific regions in the model or when it temporally occurs during training. 
These topics are discussed in more detail below in \cref{sec:local_timing}.

However, as this formulation relies on the presence of labels to begin with, it can often be challenging to quantify memorisation in settings where these are absent (e.g. generative modelling).
We, thus, turn to influence functions, which allow for such quantification to be performed efficiently in the following section.
  
\begin{tcolorbox}[colback=gray!10!white, colframe=white!50!black, title=Key Point]
Many deep neural networks have sufficient memorisation capacity to fit large datasets of completely random input-output associations.  
\end{tcolorbox}

\section{Measuring memorisation through influence estimation}
\label{sec:influence}
Influence analysis dates back to \cite{cook1977detection} and aims to offer a data-centric approach to explaining a model's predictions by apportioning credit (positive or negative) to individual training data samples.
If applied to quantify the impact that a single training data point has on the prediction of a single test sample, it is often called \textit{point-wise influence} and is equivalent to the quantity (cross-influence) introduced in \cref{eq:cross-infl}.
This canonical definition of influence (commonly known as leave-one-out (LOO) influence) is computationally expensive.
Its simple Monte Carlo estimation requires $\Omega(n)$ runs of training algorithm $A$ to convergence.
Fortunately, substantial research effort has gone into developing efficient methods for influence estimation, which will be discussed next.

As covered previously in \cref{sec:feldman}, memorisation can be quantified through \textit{self-influence}. 
Thus, all methods discussed in this section can be used as valid approaches to estimate memorisation (if applied correctly to measure a sample's influence on its own prediction).
Due to space limitations, we omit a comprehensive discussion of this topic and refer to \cite{hammoudeh2022training} for an extensive survey on influence and its estimation.

\subsection{Efficient influence estimation}
\label{sec:eff_influence}
Approaches to estimating influence can be broadly categorised into re-training-based methods and influence functions.

\textbf{Re-training-based methods} approximate \cref{eq:cross-infl} directly through simple Monte Carlo sampling.
To circumvent the aforementioned computational complexity, Feldman and Zhang \cite{feldman2020neural} proposed a \textit{sub-sampling estimator}.
This method simultaneously leaves out multiple training samples randomly (instead of just one) and records each sample's membership and the result, which are subsequently averaged to obtain a smoothed estimate of a sample's influence.
Correspondingly, the number of required training runs is drastically reduced, as every run of Algorithm $A$ now contributes to the influence computation of many samples, rather than just one.
In their work, the authors showed that this estimator is statistically consistent and that its error is bounded (with high probability) by the sub-sampling rate and the number of repetitions per randomly sampled subset \cite{feldman2020neural}.
In practice, this method allows for accurate influence estimation (self/cross-influence) in multiple hundreds or a few thousand runs on large datasets \cite{feldman2020neural, zhang2021counterfactual}.

\textbf{Influence functions} were first ported from the domain of robust statistics to deep learning through the seminal work by Koh and Liang \cite{koh2017understanding}.
Using a classical result from robust statistics \cite{cook1982residuals}, Koh and Liang showed that a change in loss (due to removal of a sample) can be approximated by through the inverse empirical Hessian (matrix of second-order derivatives w.r.t model parameters).
This means that no model re-training is required.
Intuitively, through a second-order Taylor series expansion, the effect of a small perturbation (which is assumed to be induced by removing a single sample) on the loss function can be approximated.
Hessian matrices are infeasible to compute for large neural networks, requiring memory which is square in the number of network parameters.
However, efficient approximation techniques have been proposed to tackle this issue \cite{guo2020fastif, schioppa2022scaling}, allowing for the application of influence functions even to large language models \cite{grosse2023studying}.

Re-training-based approaches, given enough computational resources, are able to provide reasonably accurate influence estimates, even for large datasets such as ImageNet \cite{feldman2020neural}.
On the other hand, influence functions have received substantial criticism \cite{bae2022if, basu2020influence, schioppa2023theoretical}, as key assumptions of the underlying theory, e.g. strong convexity and positive definiteness of the Hessian (amongst others), are not satisfied in the context of deep neural networks.
Furthermore, multiple works have shown that influence function values correlate poorly with LOO influence values \cite{bae2022if, basu2020influence}, an effect that seems to be aggravated with network size and depth. 
Thus, based on the currently available evidence, we deem re-training-based methods the preferred choice for estimating memorisation if the available computational resources allow for it.

An alternative approach to the two above-mentioned methods relies on using the so-called \say{representer theorem}, allowing the model owner to decompose the prediction into individual contributions (i.e. influences) from each data point \cite{yeh2018representer}.
This method has however been referred to as \say{overly reductive} \cite{yeh2022first} in a follow-up work by the same authors as it's limited to detecting changes in the final layer only.

\begin{tcolorbox}[colback=gray!10!white, colframe=white!50!black, title=Key Point]
  Memorisation can be measured efficiently through influence estimation techniques, but extra care is required to make sure such estimates are accurate.
\end{tcolorbox}

\subsection{Connections to data valuation}
\label{sec:valuation}
Data valuation is concerned with assigning value to data points to determine equitable monetary compensation for data owners. 
It is closely connected to the concept of influence i.e. when influence is applied to quantify the impact of a single training sample on many test samples. 
In fact, the notion of the aforementioned LOO influence is also frequently used in data valuation.
Note also that, any data valuation measure can be used as a point-wise influence measure \cite{hammoudeh2022identifying} which means that conversely any data valuation measure can be used to measure self-influence.
As influence measures were not originally designed to be equitable, substantial effort has gone into the game-theoretic analysis and improvement of data valuation/influence measures which satisfy a number of axioms aimed at improving data valuation equitability.

Arguably, the most popular valuation metric is the Shapley value \cite{ghorbani2019data} which quantifies the expected marginal contribution of individual data points while considering all possible subsets of the dataset and their interactions.
While such improved (re-training-based) data valuation/influence estimation methods have a clear advantage for equitable data owner compensation, the relevance of such game-theoretically motivated axioms for the quantification of memorisation remains unclear. 
Concretely, it has been argued by Zhang et al. \cite{zhang2021counterfactual}, that LOO influence (which does not model data point interactions), is to be regarded as the preferred metric for measuring memorisation since it allows for making \textit{causal} statements about the memorisation of a data point. 
Further, the authors argue that it's the insensitivity of LOO influence to data duplicates (a behaviour previously criticised in the context of data valuation \cite{ghorbani2019data}) which renders it a suitable metric for measuring the memorisation of rare data points.

\begin{tcolorbox}[colback=gray!10!white, colframe=white!50!black, title=Key Point]
  Game-theoretic data valuation measures, while closely connected to influence, might only be partially suitable to measuring memorisation. 
\end{tcolorbox}

\section{Quantities ostensibly related to memorisation}
\label{sec:proxies}
Beyond the techniques which can be used to directly estimate memorisation introduced in \cref{sec:feldman} (i.e. through self-influence), there exists a large number of methods which attempt to rank training samples based on their \say{importance} to the model.
Here the term importance encapsulates concepts such as difficulty of fitting, error/gradient magnitude, etc..
While, ostensibly, some of these metrics are associated with memorisation (and are often described as computationally inexpensive approximations of memorisation), we note that none of them estimate self-influence directly.

\subsection{Gradient-based influence proxies}
\label{sec:grads}
The methods introduced in this section leverage the gradient of the loss function evaluated at individual samples, either with respect to the model weights or to the inputs themselves.
Recall that the gradient is a linear/first-order approximation to the effect of a sample on the loss.
The main benefit of the techniques discussed below is that they are easy to implement and computationally efficient.

\textbf{Gradients w.r.t. model parameters}
Arguably, the most widely used gradient-related metric is its magnitude, i.e. the $\ell_2$-norm with respect to the model's parameters \cite{chen2020optimal, amiri2021convergence, lai2021oort}. 
Some works claim that the gradient norms of individual data points are proxies for memorisation (as, similarly to loss values, they tend to decrease over the course of training, highlighting that samples get \say{gradually} memorised) \cite{zhu2022isfl, katharopoulos2018not, li2021sample, xue2021toward}.
However, while higher gradient norms (similarly to higher loss values) are usually associated with difficulty when making predictions on individual data points (often correlated with samples which have higher influence), this is not a direct representation of memorisation \cite{zhu2022isfl, li2021sample}.
Moreover, even when viewing these metrics through the lens of adversarial susceptibility, there is no clear causal relationship between the amount of information memorised about individual samples; amount of information contained in these samples; and the information exposed by the model.
The gradient norm is also central to differentially private stochastic gradient descent \cite{Abadi2016} and especially in approaches employing individual privacy accounting \cite{feldman2021individual, yu2023individual, koskela2022individual}, where the gradient norm is proportional to the individual's privacy loss.
Despite that fact, and while larger gradient norms can be associated with higher attack susceptibility, this is not always observed, implying that additional factors may be involved\cite{usynin2023beyond, geiping2020inverting, balle2022reconstructing}.

We note that, techniques based on the second-order derivatives, which estimate the curvature of the loss function with respect to individual inputs (instead of weights) have also been used to measure memorisation \cite{garg2023memorization}.
Moreover, more computationally efficient techniques not requiring second-order derivatives have been developed, such as techniques which \textit{trace} the evolution of the loss function's value or the (first order) gradient through training \cite{pruthi2020estimating, hammoudeh2022identifying}.
These metrics can also be used to either A) identify samples which are more challenging for the model to learn from or B) be used to approximate the aforementioned influence functions (which are, in turn, often used to quantify memorisation).

\textbf{Gradients w.r.t. inputs}
Orthogonal to the aforementioned methods are techniques which utilise gradients with respect to the inputs.
The most notable technique in this category is the \textit{variance of gradients} technique by \cite{agarwal2022estimating}, which combines tracing the gradient values with respect to the inputs during training with the computation of their variance over the time axis. 
A higher variance is claimed to imply that the sample is more atypical and is, hence, more difficult to learn.
Thus, the authors state that samples with high variance of gradients are also more prone to memorisation.
A distinct line of work \cite{mueller2022input, hannun2021measuring} from the domain of privacy-preserving ML also used gradients (or second-order derivatives) with respect to inputs to measure privacy loss. 
These can be used to identify samples which have \say{more revealing} input features and be more susceptible to attacks on privacy, allowing the data owner to identify the privacy risks associated with individual training records. 
While privacy loss, attack susceptibility and memorisation are closely connected, these connections are, again, not causal.

Overall, many properties of training data which can be efficiently represented through the gradients (e.g. privacy loss) are often correlated to the magnitude of memorisation of that training sample.
However, these links are not causal and care should be taken when using gradient-derived metrics as approximations of memorisation.
Moreover, several works point to a connection between neural network geometry and memorisation, which we thus regard as a promising area for future research. 

\begin{tcolorbox}[colback=gray!10!white, colframe=white!50!black, title=Key Point]
  Gradient-based metrics are often used to (efficiently) measure related quantities (e.g. sample complexity), but are not direct proxies for memorisation. 
\end{tcolorbox}

\subsection{Quantifying memorisation using information theory}
The next set of techniques we discuss attempt to quantify the \say{flow of information} from data points to the parameters of a model using the tools of information theory.
A subset of these works rely on the concept of Shannon mutual information (MI) \cite{kolmogorov1956shannon}, which quantifies the amount of \say{information} which can be deduced about a variable by monitoring another variable (i.e. the amount of \say{dependence} between the variables) \cite{shwartz2022information, goldfeld2018estimating}.
However, while such methods have strong theoretical foundations, they ultimately suffer when placed in the context of ML.
This is because these approaches assume that all inputs of interest including the resulting ML model's weights are random variables, which is often not the case, as most ML approaches result only in a single set of weights.
Moreover, Shannon information assumes that computational abilities are generally unbounded.
This means that, in the view of Shannon information theory, information cannot be increased by further processing.
This stands in contrast to real-world practice, in which computation is constrained by cost or by the available time, and information is thought to be \say{extracted} from data by the ML model.

This motivated the development of usable information theory and its variant of information termed $\mathcal{V}$-information \cite{xu2020theory}.
This method can be seen as a computationally constrained version of Shannon MI. 
In contrast to Shannon MI, $\mathcal{V}$-information measures the \say{usable} information contained in the data, which can be extracted by functions (e.g. ML models) in a specific family (denoted $\mathcal{V}$).
This interpretation has been applied to quantify the difficulty of datasets \cite{ethayarajh2022understanding} and to localise information leakage to specific components (layers) of ML models \cite{mo2021quantifying}.
The authors of the latter work also correlated $\mathcal{V}$-information with Jacobian sensitivity, which corresponds to the norm of the gradient with respect to the model's input, linking this study to the techniques discussed above.
They deduce that there are two \say{types} of useful information the model can memorise: \textit{input} and \textit{latent} information.
The former represents the model's ability to correctly recall the input data under reconstruction attacks \cite{zhao2019inferring}, while the latter corresponds to information about properties of the data (which can be exposed through attribute inference attacks, discussed below).

Orthogonal to the domain of usable information theory are approaches which attempt to enable the computation of information-theoretic quantities in high-dimensional ML models.
Notable among these is the technique of estimating e.g. MI through randomised low-dimensional projections, pioneered in \cite{goldfeld2021sliced} and termed \textit{sliced} MI.
Sliced MI was used in \cite{wongso2023using} to measure memorisation and captures the notion of average usable information (as it averages the MI for all random projections) instead of the largest usable information.
We contend that the relationship between the difficulty of learning, the complexity of the representations at hand and the memorisation properties of the network are still a nascent area of research of high potential impact.

Another noteworthy approach was proposed by \cite{harutyunyan2021estimating} and is termed \textit{smooth unique information}.
This formulation considers the distinction between what data the weights of the model contain against what the model actually learns.
Similarly to influence functions, authors consider how a single included/excluded instance can influence the training of the model and measure it using a smoothed KL-divergence between the two models (concretely: its expectation over the label distribution).
Authors conclude that removing highly informative samples identified by their method results in a stronger performance degradation compared to removing the same quantity of uninformative/random samples instead.
These findings again highlight the divide between studies on informativeness of individual samples and the extent to which they are prone to memorisation.
While this work studies the effect of including or excluding highly informative (in a manner of speaking -- influential) samples on generalisation performance, no analysis is performed in terms of the influence of training samples on themselves.

\begin{tcolorbox}[colback=gray!10!white, colframe=white!50!black, title=Key Point]
  There is no unified perspective on how memorisation can be localised in the information-theory domain, but prior works agree that A) memorisation can be localised and B) different concepts are memorised in different parts of the model.
\end{tcolorbox}

\subsection{Measuring sample difficulty}
The concept of sample difficulty is often closely related to memorisation (e.g. \say{difficult} samples can be more prone to memorisation), but is arguably ill-defined.
One can intuitively interpret a difficult sample to be one which is fit poorly by the model.
However, the reasons for such poor performance can vary: the sample can contain \say{rare} features \cite{agarwal2022estimating} or have poor quality \cite{usynin2023leveraging}; alternatively the sample can come from a data-generating distribution which is mismatched to the task at hand \cite{chen2021synthetic} or from the correct data-generating distribution, albeit from a low-density region.
A number of methods were proposed to identify at which point different \say{concepts} are learned from individual features of a sample, showing that those that correspond to \say{easier} concepts are identified much earlier in the model \cite{baldock2021deep}.
Alternatively, the authors of \cite{garg2023samples} rely on the curvature of the loss function to assess how \say{clean} individual samples are (where higher cleanness signifies a stronger representative of a class).
However, similarly to other works on example difficulty, the notions of \say{cleanness} are closely intertwined with samples being \say{typical} representatives, which can have different interpretations based on the input modality (i.e. defining a typical cat image vs. a typical job description). 
Notably, this method through an approximation of the Hessian trace produces consistent results across different architectures and random seeds (which was a notable limitation of similar approaches \cite{agarwal2022estimating, pruthi2020estimating}).

Arguably, the best example of the challenges associated with the interpretation of \say{sample difficulty} is discussed in \cite{carlini2019distribution}, as this work employs five separate metrics in order to identify the connections between individual image characteristics (for instance, adversarial robustness on these images) and its \say{difficulty}.
While authors conclude that most of these metrics are correlated amongst each other, there is still no causal relationship between e.g. holdout accuracy and \say{difficulty} of a sample.
Another work of \cite{harutyunyan2021estimating} shows that example difficulty can be estimated by observing the behaviour of the loss values on individual samples on their \say{corresponding} test-time points.
What this definition proposes, is to disentangle concepts of train-time performance and example difficulty, and to, instead, consider related data points, that the model has not encountered during the learning process.

Finally, another noteworthy work which tries to approximate example difficulty was proposed by \cite{ethayarajh2022understanding} using $\mathcal{V}$-usable information (which is a computationally constrained version of Shannon entropy). 
What we are interested in here is the similarity of this definition to the notion of sample influence.
Here authors estimate the difficulty of a sample as a model's predictive performance on two data distributions, one of which contains a record of interest and one does not.
This metric describes the difficulty of an entire dataset, given a model $\mathcal{V}$, but authors also proposed pointwise $\mathcal{V}$-information (PVI), which can be used to describe the difficulty of individual samples, similarly to the works we discussed above.
One noteworthy conclusion of this work is that PVI, can serve as a context-agnostic measure of \say{difficulty} across different models and modalities if using it as a threshold (i.e the PVI value after which samples start to be misclassified and are hence considered to be \say{difficult}).
These methods establish an indirect connection between difficulty and memorisation: Difficult samples can often be those that fall on the tails of the data distribution and are, hence, more prone to memorisation.
However, reasoning over sample difficulty is not straightforward, because while difficult samples may be highly self-influential (e.g. mislabelled data), they do not necessarily have to be (e.g. poorly acquired images).

Overall, it is speculated by several works that the identification of \say{difficult} samples is similar to the identification of samples which are likely to be memorised.
These methods can even be perceived as proxies to the aforementioned influence functions using only first-order information.
While the definition of \say{sample difficulty} has not been unified yet across various works, most of the formulations that we discussed base their definitions of the concepts of \say{atypicality}, making detection of such samples a challenging, yet an important task across any input domain.

\begin{tcolorbox}[colback=gray!10!white, colframe=white!50!black, title=Key Point]
Samples can be difficult for many reasons, but samples which are deemed to be more difficult are often memorised more.
\end{tcolorbox}

\section{Localisation and timing of memorisation}
\label{sec:local_timing}
Two intuitive question which arises from the discussion above is: 
\begin{enumerate}
    \item Where does memorisation occur in the model?
    \item When does memorisation occur during training?
\end{enumerate}

\begin{figure*}[!htp]
    \centering
    \begin{subfigure}{0.4\linewidth}
    \centering
    \includegraphics[width=\linewidth]{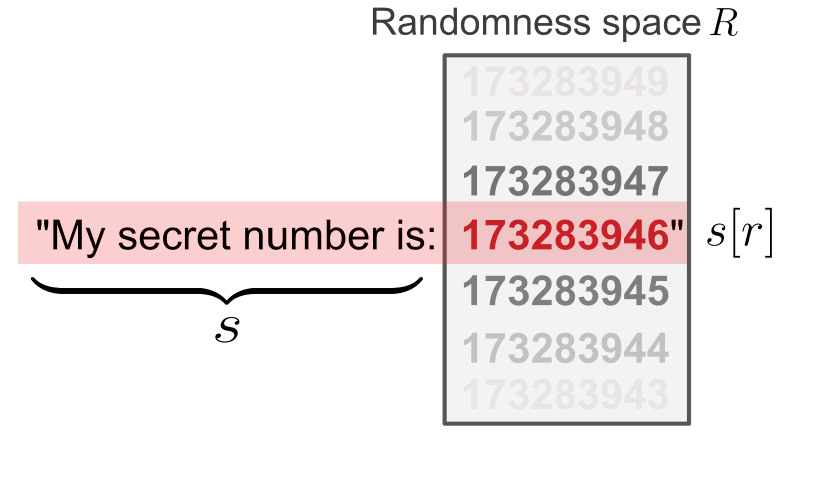}
    \caption{}
    \label{fig:canary_a}
    \end{subfigure}
    \vspace{5mm}
    \begin{subfigure}{0.45\linewidth}
    \centering
    \includegraphics[width=\linewidth]{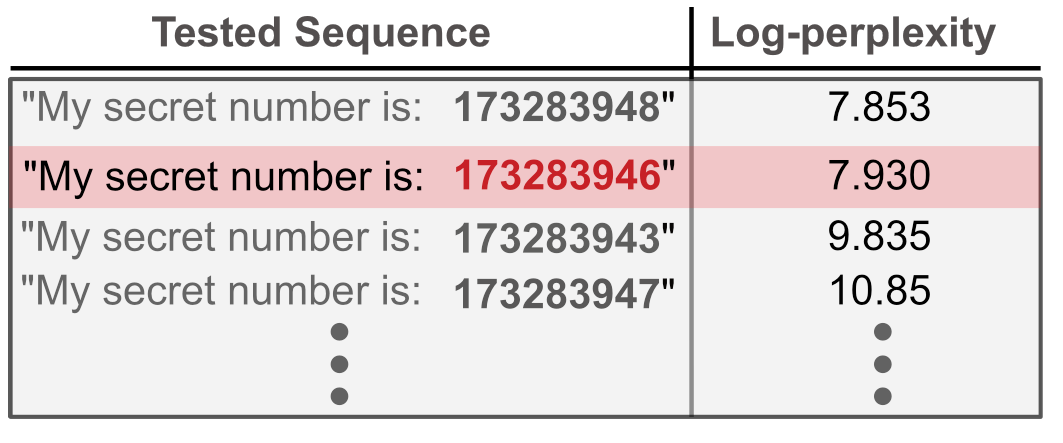}
    \caption{}
    \label{fig:canary_b}
    \end{subfigure}
    \caption{Example workflow of measuring memorisation through an inserted canary (shown in \textcolor{BrickRed}{red}). First, a canary is generated by combining some pre-defined structure $s$ and some randomness $r$ drawn from some randomness space $R$ (e.g all possible nine-digit numbers) (a). After inserting the canary into the training dataset, the ability of the model to reproduce the canary is computed as its \textit{exposure}. This measure is determined by $|R|$ (the size of $R$) and the canary's index (rank) in table (b), ordered by log-perplexity and containing all other possible canaries from $R$ (that could have been inserted but were not).}
    \label{fig:canary}
\end{figure*}

To reason over the localisation of memorisation, we first need to consider the localisation of \textit{learning} in ML models.
One of the fundamental works presented by \cite{bau2017network} shows that individual concepts that are included in a data point (e.g. texture or color of an object on an image) are learnt by different parts of a neural network.
Authors demonstrate that during the learning procedure, different parts of the model become \say{responsible} for learning individual concepts contained in the input. 
Thus, it is possible to identify individual neurons associated with specific concepts.

As learning of individual concepts can be localised to different parts of the model, it was previously established that this also holds for memorisation.
It was shown in the works of \cite{mo2021quantifying, baldock2021deep, maini2023can} that this is indeed the case and that as part of the learning process, memorisation can also be associated with specific parts of a model.
Specifically, authors of \cite{baldock2021deep} present \textit{prediction depth}, which is the earliest layer in a model based on the representations of which it is possible to make a correct prediction on the input. 
Authors determined that \say{easier} concepts (i.e. those associated with generalisation) are learnt in the earlier layers, whereas the more complex ones are learnt in the last layers of the model.
The work of \cite{stephenson2021geometry} further shows that memorisation of atypical (mostly mislabelled) samples can also be localised to the last layers of the model, where most mispredictions also occur in the early stages of training.
It is of note that the work of \cite{mo2021quantifying} shows a similar result but through the lens of gradient information leakage, which we have previously discussed in \cref{sec:proxies}.
The fact that memorisation can be localised was further supported by the work of \cite{maini2023can}.
However, authors of \cite{maini2023can} show that contrary to the results of \cite{baldock2021deep}, memorisation, while localised to specific parts of the model, is \textbf{not} associated with the final layers.
They instead show that memorisation is indeed localised to specific neurons, but these are often distributed across multiple layers.
Moreover, the authors of \cite{wongso2023using} conclude that learning patterns can be different based on the architecture and the depth of the model: convolutional models exhibit more \say{learning} in the deeper convolutional layers, similar to the discussion above.
In turn, multi-layer perceptron-based models exhibit an increase in the amount of useful information the model learns approximately linearly (with an increase in the depth of the layer).
These findings highlight that learning (and memorisation) is not only inconsistent across different model types (i.e. different models can extract different useful information), but also that even individual \textit{layers} in a model learn differently from the same data based on how deep they are located (i.e. learning and memorisation can be localised to specific parts of the model).

In regards to the question of timing, authors of \cite{arpit2017closer} show that memorisation is prevalent at specific stages of training.
Concretely the authors showed that models tend to start learning \say{simple} patterns first, thus showing that generalisation can often occur before memorisation (further supported by \cite{kishida2019empirical}).
Additionally, the work of \cite{paul2021deep} shows that it is also possible to determine the influential samples early on and use these to guide the training process, by removing samples of low influence from the training dataset.
Therefore, it is also possible to not just link memorisation to the training phase but to also employ the training data to force the model to change its memorisation pattern (i.e. which samples get memorised). 
This mirrors a finding by \cite{baldock2021deep}, who relate memorisation to \textit{prediction depth}, i.e. the depth of the layer at which representations which effectively determine the network's prediction on a specific sample are formed.
This is intended to highlight that a higher difficulty of forming a defining representation in a network is associated with a higher probability of memorisation.

Overall, although there is no clear consensus regarding the specifics, evidence from multiple prior works supports that memorisation is a process that can be localised both spatially and temporally.

\begin{tcolorbox}[colback=gray!10!white, colframe=white!50!black, title=Key Point]
Memorisation is a process that can be localised both spatially and temporally.
\end{tcolorbox}

\section{Inducing deliberate memorisation}
For many learning settings and in particular, for generative models, it is often very difficult to conceptualise what the model memorises (i.e. does producing an output \say{similar} to the training record count as memorisation?)
One intuitive method of evaluating this involves deliberately inserting data samples which are \say{expected} to be memorised by the model due to their \say{atypicality}. 
The capacity for memorisation is then measured with respect to how well the model can reproduce these inputs as its output (i.e. for generative models) or how performance on these data points compares to performance on the rest of the dataset (e.g. for supervised learning).

\subsection{Canary memorisation}
\label{sec:canary}

The, arguably, most widely used approach for inducing (and empirically quantifying) memorisation was proposed by Carlini et al. \cite{carlini2019secret} and has since been employed in many machine learning contexts \cite{thakkar2020understanding, carlini2021extracting, tirumala2022memorization, lee2021deduplicating}, particularly in the field of language processing. 
In essence, the authors propose to measure how much a generative language model can memorise by purposefully inserting crafted data samples (canaries) into the training dataset.
These crafted samples are designed to require memorisation.
The connection to the long-tail definition of \cite{feldman2020does} is that such canaries are constructed to resemble data from low-density regions of the training data distribution.
These samples are (often informally) called anomalous, outliers or atypical. 
Note that in supervised learning, a similar effect can be achieved through deliberately mislabelling the example.
The \say{exposure} of the canaries is measured and used as a proxy for the memorisation of other atypical samples. 

To define exposure, we first discuss the notion of \textit{log-perplexity} it is based on. 
Intuitively, log-perplexity measures how \say{expected} a given sample is (or how much the model is \say{surprised} by this sample) and is defined as follows: 
\begin{align}
    \mathrm{perp}(x, f) = & - \log_{2}p(x_1,...,x_m | f) \\ = & \sum^{m}_{i=2}(-\log_{2} p(x_i | f(x_1,...x_{i-1}))) \notag
\end{align}
where $f$ is the model, and $x$ is the input sequence of length $m$.
Here and below, we use $p$ in a slight abuse of notation to denote both probabilities and likelihoods. 

To generate a canary ($s[r]$), we combine a predefined, fixed structure $s$ and some randomness $r$ drawn from some predefined randomness space $R$ (e.g. the space of nine-digit numbers, see \cref{fig:canary_a}).
This generated canary is then inserted into the training dataset.
After model training, we now wish to compare the perplexity of our inserted canary to all other possible canaries that we could have inserted.
Perplexity, however, is a relative measure and highly dependent on the specific training setup, including model architecture, dataset composition and the specific application. 
Thus, Carlini et al. \cite{carlini2019secret} proposed a \say{relative} notion of perplexity called \textit{rank}, i.e. the index of a canary in the relative ordering of perplexities across all possible canaries (see \cref{fig:canary_b}).

Exposure, in turn, is then simply determined by the size of the chosen randomness space and the canary's rank.
Formally it is defined as follows:
\begin{equation}
    \label{eq:exposure}
    \texttt{exposure}_f (s[r]) = \log_{2}|R| - \log_{2}\texttt{rank}_f (s[r]),
\end{equation}
where $|R|$ is the size of the randomness space from which the canary $s[r]$ was generated, thus making exposure a positive value.

The exposure definition has been successfully transferred to other machine learning modalities, such as computer vision in \cite{hartley2022measuring}, but due to the complexity of canary generation in other domains, most follow-up works in the area concentrated on language models (LMs) instead \cite{thakkar2020understanding, carlini2021extracting, tirumala2022memorization}. 
However, canary-based memorisation definitions are significantly less generalisable than the one discussed in \cref{sec:feldman}, as they (1) require the data owner to generate canaries (which are specific to their setting) and (2) are particularly well-suited to language settings, compared to other generative modelling tasks.

Overall, canary-based approaches have seen popularity due to their simplicity of implementation and high applicability to generative ML tasks \cite{liang2023code, nijkamp2022codegen, taylor2022galactica, alayrac2022flamingo, wei2022emergent}.
One important finding that authors of \cite{carlini2019secret,carlini2022quantifying, carlini2023extracting}) report, is that exposure seems to increase drastically with the number of times a canary is seen during training. 
The authors thus suggest that if an atypical sample is included many times in the training dataset, it is much more likely to be memorised.
This is in contrast to follow-up work \cite{zhang2021counterfactual}, which has argued that such a conclusion is ill-guided, as extraction/generation-based memorisation quantification methods (like canaries) are biased towards identifying frequently occurring samples \cite{lee2021deduplicating, zhang2021counterfactual}.
In fact, the \cite{zhang2021counterfactual} presents contrary evidence to the prior claim: using re-training-based influence estimators (from \cref{sec:eff_influence}) the authors show that memorisation estimates tend to actually decrease for data points that are repeated many times.
This finding thus raises the question of the validity of memorisation and privacy risk conclusions drawn from extraction-based memorisation experiments, as the memorisation of frequently occurring sentences (or data points) potentially poses little privacy risk compared to rare sentences.
This is because frequently occurring sentences such as commonly known facts typically pose little privacy concern compared to e.g. personally identifiable information which only occurs once.

\begin{tcolorbox}[colback=gray!10!white, colframe=white!50!black, title=Key Point]
It is possible to measure unintended memorisation by inserting samples into the training dataset which are deliberately crafted to be more prone to memorisation and measuring how likely they are to be regurgitated by the model.
However, the conclusions of such experiments should be interpreted with care.
\end{tcolorbox}

\subsection{Memorisation and adversarial samples}
The aforementioned canaries are deliberately crafted to resemble samples from the low-density region of the data distribution.
They are, however, still valid (albeit atypical) data points and can be used to train a well-generalised model.
The same concepts are exploited in a line of work on adversarial samples, which generate data points from the low-density region of the data distribution, but are crafted to degrade the performance of the model.
These, unlike canaries, are applicable to both the discriminative and the generative settings (and are assumed to have high negative cross-influence in addition to high self-influence).

In many cases, a very small proportion of such samples ($<1\%$) suffice to severely harm the utility of the trained model \cite{chobola2022membership, zhou2021deep}.
Adversarial data points are generated to be atypical through either incorrect labelling, embedding of features which are associated with a different class, addition of imperceptible noise etc. \cite{usynin2021adversarial}.
As a result, these samples can often be used to manipulate the behaviour of the model as these are A) more likely to be memorised and B) only require a small perturbation, making these attacks difficult to detect.
Adversarial samples can also be used to aid attacks on ML models, aiming to extract the information that the model has memorised (particularly for underrepresented samples, which were shown to be more vulnerable in \cite{shokri2017membership}) about individual data points, which it would not expose otherwise \cite{tramer2022truth, carlini2022privacy, chobola2022membership, bagdasaryan2021blind}.
For a more in-depth discussion on how these adversarial data can be generated in different learning settings, we refer to \cite{tian2022comprehensive}.

\begin{tcolorbox}[colback=gray!10!white, colframe=white!50!black, title=Key Point]
Adversarial samples are made to be artificially atypical and hence have a higher influence on the model (malicious or otherwise).
\end{tcolorbox}

\section{Privacy implications of memorisation}
\label{sec:implications}
Memorisation is not a process that occurs in a vacuum: it likely affects the privacy of individuals, whose data is used to train a ML model.
To assess the extent to which this phenomenon can harm these individuals, we discuss the implications of memorisation through the lens of data privacy.
We are particularly interested in answering these questions over the course of this discussion:
\begin{itemize}
    \item What are the implications of memorisation on privacy of the individuals and which definition(s) can we use to estimate these?
    \item What methods exist to extract the information memorised by a model?
    \item How can memorisation affect privacy of generative models?
\end{itemize}

\subsection{Memorisation and privacy attacks}
\label{sec:attacks}
Models which have memorised much of the data they have been trained on can often be perceived as more \say{dangerous}, as in many cases it is possible to extract the information that has been memorised.
While attacks can be executed in both the discriminative and the generative settings, we are particularly interested in discriminative models.
For these extraction of memorised information often takes the form of attacks on privacy, as they are not designed to output the data they were trained on.
We discuss which \say{individual} traits of ML training can be exploited to extract the information the model memorised about individual samples (similarly to how many of \textbf{the same} metrics can be used to identify influential or atypical samples). 
Concretely, we discuss three major privacy attack methods, namely: membership inference, attribute inference and data reconstruction (often referred to as data inference or model inversion) \cite{usynin2021adversarial}. 

\textbf{Inference attacks}
One of the more prominent privacy attacks is termed a \textit{membership inference attack} (MIA) \cite{shokri2017membership} and it identifies if the model's training dataset contains a specific point of interest. 
This attack has a flexible threat model and can be employed against shared white-box models \cite{sablayrolles2019white} or API-only black-box models \cite{choquette2021label} with high effectiveness.
Similarly, MIAs can successfully target both the discriminative and the generative models \cite{chen2020gan, liu2019performing, hilprecht2019monte}, making them a versatile context-agnostic \say{unintended memorisation} auditing tool.
There is growing evidence to support that high memorisation is associated with an increased MIA vulnerability, although it seems that memorisation is not necessary \textbf{and} sufficient but just sufficient (i.e. where memorisation leads to being vulnerable to MIAs) and that there are other reasons for vulnerability beyond memorisation \cite{choi2023train, carlini2022privacy}.

\begin{figure*}[h]
    \centering
    \includegraphics[width=0.85\linewidth]{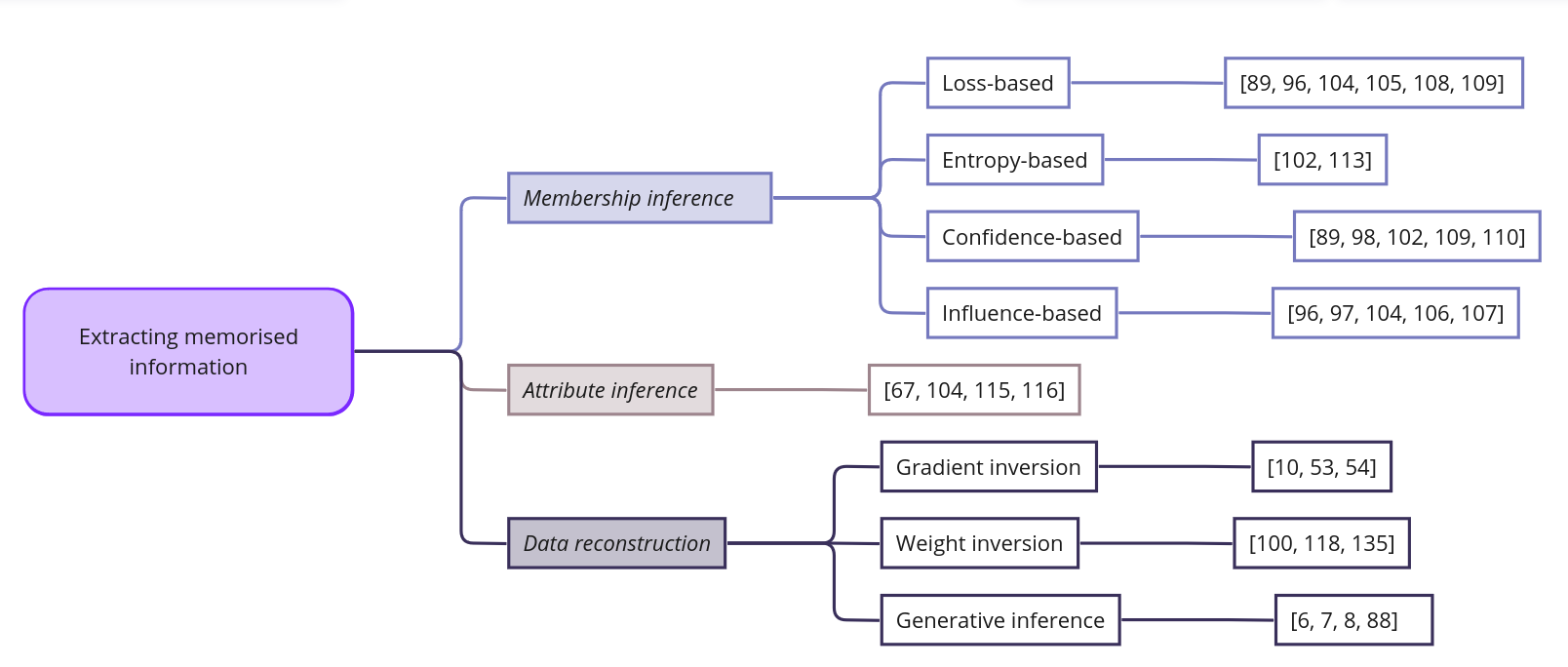}
    \caption{Brief overview of various attacks on privacy and the information they exploit.}
    \label{fig:attacks}
\end{figure*}

It was previously shown that MIAs are particularly effective against models with larger generalisation gaps \cite{shokri2017membership, usynin2022zen, truex2019demystifying, salem2018ml}. 
For instance, a suite of attacks all of which rely on comparing the loss values on previously seen and unseen samples \cite{chobola2022membership, carlini2022membership, yeom2018privacy, jayaraman2020revisiting} naturally benefit from lower loss magnitudes on the data the model was trained on.
However, it is important to note that relying on model loss, while improving the naive prediction-based approaches (e.g. \cite{yeom2018privacy, choquette2021label, sablayrolles2019white, irolla2019demystifying, bentley2020quantifying}), are typically uncalibrated and represent a non-membership test (i.e. are mostly effective at determining who was \textbf{not} a member rather than who was one) \cite{li2022privacy}.
Similarly, any proxy metric, such as the confidence scores, can be used to identify whether a training record was previously \say{seen} by the model \cite{shokri2017membership, salem2018ml, watson2021importance, gu2022cs}. 
However, it was previously shown in \cite{yeom2018privacy} that overfitting is \textbf{not} a prerequisite for a successful attack.
Moreover, when one considers attacks on generative models (e.g. those described in \cite{chen2020gan, hayes2017logan, hilprecht2019monte}), it is not clear what overfitting even implies.

Therefore, works such as \cite{jayaraman2019evaluating, song2021systematic} attempt to make a more concrete connection between how memorisation of data points affects their inference attack susceptibility (based on the notion of memorisation presented in \cref{sec:canary}).
This formulation permits a direct evaluation of how well individual samples can be A) memorised and B) inferred by a MIA adversary. 
So while a direct causal relationship between memorisation and susceptibility to MIA has not yet been established (and although it seems likely), there are still two noteworthy observations.
Firstly, atypical samples which are more likely to be memorised, are also more susceptible to inference attacks, showing that there is a connection between how much can be memorised by the model and how much can then be \say{exposed} by the model \cite{mo2021quantifying}.
Secondly, similarly to the formulations of memorisation, the extent of MIA's success can also depend on memorisation and generalisation simultaneously, showing that these notions are complementary.

Another attack, which allows the adversary to determine individual features or attributes of their victim is termed as \textit{attribute inference} \cite{kosinski2013private}.
For instance, it was previously shown that attribute inference attackers are capable of inferring protected demographic information from parameter updates alone \cite{feng2021attribute, jourdan2021privacy}.
As previously discussed in \cite{mo2021quantifying}, the risk of this attack can directly depend on the amount of information that is \say{embedded} in the trained models (more so in deeper layers \cite{maini2023can}).
Therefore works such as \cite{thomas2020investigating, jourdan2021privacy} hypothesise that similarly to MIAs, this attack also benefits from memorisation of individual samples and is particularly effective against data points of high influence.

\textbf{Model inversion}
The last type of privacy attacks that we discuss is model inversion attacks, which aim to extract the training data given some representation of the model (e.g. consecutive training snapshots \cite{usynin2022zen}, final model \cite{haim2022reconstructing}, gradient updates \cite{zhu2019deep, geiping2020inverting, usynin2023beyond}). 
The unique trait of these attacks is the fact that they can be used against \textit{different} representations of the same model in order to extract the training data. 

When quantifying memorisation under these attacks it is important to clearly disentangle the notion of memorisation and \say{leakage} (or exposure).
Memorisation is the amount of information the model can \textit{store} about individual (usually rare) samples, whereas leakage is the amount of information that the model can \textit{reproduce} when queried.
Therefore, while many of these attacks show \say{how much models leak}, these leakages can often A) represent the information about a class of input samples rather than individual samples (e.g. reconstruction in \cite{fredrikson2015model}) and B) not capture all of the information memorised by the model (e.g. when extracting an image, a lot of pixel variation is possible, making it unclear what exactly was memorised by the model, while still showing \say{some} sensitive attributes \cite{usynin2023beyond}).
These attacks typically exploit phenomena that are similar, but otherwise disjoint from memorisation (which we discussed in \cref{sec:proxies}).
Notions such as $\mathcal{V}$-information \cite{mo2021quantifying} can be employed to identify the amount of useful data contained in the shared gradient, while a more direct metric (e.g. canary-based approaches \cite{carlini2019secret}) could be used to quantify the amount of memorisation occurring over training that gets exploited by an inversion attacker.

The aforementioned attacks (in particular MIAs) can be used to \say{audit} the model and to empirically identify information it memorised. 
This was previously manifested in A) works which verify is a specific training sample was memorised during model training \cite{jayaraman2019evaluating, song2021systematic} and B) works validating the bounds on how much memorisation can occur with respect to specific samples \cite{ye2022enhanced, nasr2021adversary}.
The former allows privacy-conscious individuals to identify if their data was likely used (and memorised) by the model during training.
The latter can help both the data and the model owners to establish how much information can be both memorised and exposed once a trained model is made public (and is discussed in more detail in \cref{sec:dp}) to introduce more realistic privacy bounds.

Fundamentally, as new attacks are developed, (particularly those that only require access to the trained model \cite{haim2022reconstructing}), identifying which \textit{specific} information contributed to a successful attack becomes more challenging (and whether it is related to memorisation at all).

Finally, it is important to discuss how memorisation (as well as attack susceptibility) of one record can directly affect another record.
We have previously established that data points which are more memorised (and have a higher influence on the model) are typically associated with being more susceptible to privacy attacks.
Thus, intuitively we may deduce that removing these records would reduce the amount of memorisation a model can experience and, hence, make the entire settings more robust against privacy attacks.
Authors of \cite{carlini2022privacy} discovered that this, in fact, does not seem to be the case.
They termed this phenomenon as \textit{the privacy onion effect}, which states that should a sensitive record be removed from the training dataset, the model can relatively quickly start memorising other records more in order to fill the information that becomes missing upon an exclusion of the record of interest. 
Hence, authors given an \say{onion} analogy: when the top privacy-sensitive layer of data points gets removed (i.e. a set of records which are most susceptible to privacy attacks), the layer that follows is becomes more exposed and, thus, becomes more susceptible to the same attacks.
This phenomenon directly links to the rest of our work: the amount of memorisation that the model experiences with respect to a given data point is \textbf{relative} i.e. all other things being equal, the rest of the data used to train this model (and in particular its distribution) can significantly affect how much model memorises with respect to this point.
It is also, to the best of our knowledge, one of the few works to identify a direct relationship between exposure, attack susceptibility and memorisation, further proving that all of these are distinct, yet complimentary processes.
Note also that --by design-- auditing measures \textit{leakage} from the model and not memorisation; it thus stands to reason that auditing surfaces the points on the \say{outer layers of the onion}, i.e. the most susceptible points, rather than the most memorised ones.

Overall, identification of the risks associated with model memorisation is a non-trivial task, which relies on a large number of methods that we have previously discussed. 
More importantly as underrepresented populations are much more likely to be memorised \cite{feldman2020neural} and are often more susceptible to attacks \cite{kulynych2019disparate}, causal reasoning over the privacy risks associated with memorisation is a very important direction for development of secure AI systems.

\begin{tcolorbox}[colback=gray!10!white, colframe=white!50!black, title=Key Point]
Increasing evidence suggests that privacy attacks on ML are more effective against samples which are memorised more.
\end{tcolorbox}

\subsection{On memorisation in generative models}
As generative ML becomes more prevalent \cite{liang2023code, nijkamp2022codegen, taylor2022galactica, alayrac2022flamingo, wei2022emergent}, we find that the existing formulations of generative memorisation (and even more so - generalisation) have very different privacy implications compared to the discriminative settings.
The question which naturally arises when discussing generative models is: as new data is generated based on the previously seen one, can the model compromise privacy of an individual by generating data close to the one they shared during training?
There exists a large number of prior studies on the privacy of generative models \cite{liu2019ppgan, triastcyn2020federated, wu2019generalization, carlini2021extracting, carlini2019secret}, one of which we have previously discussed in detail in \cref{sec:canary}. 
The main takeaway from this discussion is that for individuals (and their data) which fall on the tails of the data distribution (i.e. of high influence), generative models can present a greater privacy concern, as their data is more likely to be memorised (particularly in language-based models) \cite{carlini2019secret, carlini2021extracting, ippolitopreventing}. 

The question which remains unanswered concerns the metrics which can be used to measure this unintended memorisation.
While it is sometimes possible to generate the \textbf{exact} training data used to train language models (e.g. in \cref{fig:canary}), how could one measure the same notion in a domain, where such one-to-one mapping is unlikely and where data can be generated with \say{similar enough} private features corresponding to the training data? 
Moreover, authors of \cite{ippolitopreventing} and \cite{lee2021deduplicating} argue that even for LLMs, there is no straightforward way to describe memorisation of sensitive data (as the models are capable of producing outputs, which are syntactically different, but semantically identical, making \say{verbatim memorisation} a poor privacy metric).
Similar conclusions were presented by \cite{carlini2023extracting} for images and it was termed as the \textit{eidetic} memorisation (closely linked to the concept of \say{episodic} memorisation from \cite{zhang2021counterfactual}), which refers to model's ability to output an image it has only seen once (or $k$ times for a $k$-eidetic memorisation, where $k$ is a small integer).
In order to quantify memorisation in this domain, authors proposed to use a threshold $\delta$, which measures the distance between an image and similarly-looking neighbours, alleviating the need to rely on an exact image reconstruction.

However, one may argue that since pixels (on their own) are independent, but the features within natural images are not, without further contextualisation such metrics are meaningless.
The work of \cite{fernandez2023privacy} tries to address this issue by using a similarity metric (proposed in \cite{packhauser2022deep}) based on the re-identification ratio: can the adversary identify the private data of a specific individual based on the image generated by the adversary?
Another method was previously proposed by \cite{kuppa2021towards} and here authors propose to measure generative sample memorisation through susceptibility of individuals whose data is used to train the model to privacy attacks, in particular membership inference.
We discuss how memorisation is related to information leakage in \cref{sec:attacks} in more detail, but as we can already see it is difficult to disentangle the two and measure the degree of memorisation directly.
Therefore, we conclude that while in discriminative settings there exists a number of methods which aim to identify specific features or samples the model \say{memorises}, in the context of generative modelling, these are very challenging to contextualise.

\begin{tcolorbox}[colback=gray!10!white, colframe=white!50!black, title=Key Point]
Memorisation in generative models can be perceived either through self-influence or as the ability of a model to produce outputs which are similar to the input data.
\end{tcolorbox}

\section{Preventing and reversing memorisation}
As information can be memorised by the model, it raises the question whether it can also be \say{unmemorised}.
This process can occur naturally over the training process and it can be induced manually by the data owner.
Moreover, over the course of training it is also possible to bound the amount of information that the model can memorise through the use of differentially private training.
In this section we discuss these phenomena and outline the processes which are responsible for reversal or reduction of memorisation.

\subsection{Spontaneous reversal of memorisation}
While some data points are more prone to memorisation, the information contained in the model about these memorised samples often changes over time (due to model's limited memorisation capacity). 
This effect is often termed as \say{forgetting}, which intuitively means that if the model does not encounter a previously seen training point during training (i.e. if it \say{disappears} from the training set), the performance of that model on this specific training point would degrade over time \cite{toneva2018empirical}.
The main question to consider is: \say{is forgetting the opposite of memorisation?}
Intuitively, the answer is affirmative, as when the model forgets the data point, model's performance on it degrades.
This effect can be particularly profound for the data points which fall on the tails of the distributions (e.g. mislabelled data points), with \cite{jagielski2022measuring} showing that data which is more likely to be memorised under \cite{feldman2020neural} definition is also more likely to be forgotten.

Authors of \cite{jagielski2022measuring} argue that similarly to the conclusions of \cite{feldman2020does} that ML models \textit{have to} memorise, they also \textit{have to} forget. 
This work, again, shows that the samples which are most vulnerable to this phenomenon are the atypical and the duplicated samples. 
This notion has different implications to \say{catastrophic forgetting}: where the entire sub-distribution of data (e.g. an entire class) can be forgotten, which is not guaranteed to directly affect the previously memorised samples (but rather some population of the training set, which \textit{could have been} memorised).
This phenomenon can be measured using the concept of adversarial advantage, which compares the performance of membership-based attacks on the samples of interest at different stages of training. 
Moreover, the issue of forgetting captures another important question that we have previously outlined: when \say{forgetting}, one is required to define what is that needs to be \say{forgotten} (i.e. what has previously been memorised)?
In \cite{tirumala2022memorization}, for instance, this concept captures verbatim memorisation, which authors of \cite{jagielski2022measuring} argue does not represent a more general definition of memorisation. 
Similarly to other works on memorisation in language models, they craft uniquely identifiable samples (i.e. canaries from \cite{carlini2019secret}) to approximate the degree of memorisation that remains on specific training records over the course of training.

\begin{tcolorbox}[colback=gray!10!white, colframe=white!50!black, title=Key Point]
Samples can be forgotten over the course of training if they are not encountered again. 
\end{tcolorbox}

\subsection{Induced reversal of memorisation}
Finally, we briefly mention another area which is closely linked to both notions of memorisation and forgetting: machine unlearning \cite{bourtoule2021machine}. 
Unlearning is a set of different techniques, which given a training record(s) modifies the model in order to no longer include the contributions based on that record(s). 
It is of note that forgetting \textit{can} be targeted as presented in \cite{zhou2022fortuitous}, showing that these phenomena are related.
The difficulty of unlearning depends on a variety of factors, including the current state of the model (i.e. if it is currently training or if the model has already been trained), the features of the data record(s) as well as the remaining data points.
Forgetting (when compared to unlearning) occurs naturally during training (it is often not possible to predict which specific records will be forgotten \cite{toneva2018empirical}), whereas unlearning is a set of techniques designed to remove the contributions of specific data point(s).
The final factor we outlined bears additional discussion. 

As above, when performing unlearning, one tries to disentangle and remove the contribution of one or more data record. 
If this data point is \say{simple} or its class is over-represented, then the removal of its contribution is unlikely to have a large effect on the rest of the population (e.g. their attack susceptibility). 
However, if the record(s) comes from the tail end of the data distribution, its unlearning can negatively affect privacy of those, whose data remains in the dataset. 
We have previously discussed this phenomenon, namely the \say{the privacy onion effect} \cite{carlini2022privacy}, which shows that by deliberately removing the record (and, thus, its contribution) which was deemed to be more susceptible to attacks, other samples can become more prone to inference.
Thus, we, again, observe that memorisation (and un-memorisation) is indeed relative to the data surrounding a highly memorised sample. 
As a result, this opens up a number of additional questions, such as \say{Does un-memorising also mean un-generalising? If yes, is it ethical to withdraw ones data if this will deteriorate the generalisation performance on other samples?}. 
These questions must to be deliberated by the ML community in order to devise a unified strategy for effective model training.

\begin{tcolorbox}[colback=gray!10!white, colframe=white!50!black, title=Key Point]
Forgetting can be artificially induced to \say{unmemorise} individual data points.
\end{tcolorbox}

\subsection{Bounding memorisation with differential privacy}
\label{sec:dp}
Differential privacy (DP) \cite{dpbook}, as briefly discussed previously, is the canonical privacy definition for statistical data processing and ML. 
Intuitively, DP can be perceived as a property of an algorithm, making it approximately invariant to an inclusion/exclusion of a single data point.
More formally: The result of a computation (e.g. model training) is said to be \textit{differentially private} if a probabilistic stability notion over neighbouring datasets is satisfied.
Concretely, in the context of ML (and our previously introduced notation), this stability notion requires that a specific model $f'$ (or set of weights) has a similar likelihood under $A(S)$ and $A(S\backslash i)$.
This means that the value the random variable $f$ takes, cannot change much no matter which data point we remove from the training dataset.
More formally, random Algorithm $A$ satisfies $\varepsilon$-DP when:
\begin{equation}
    p\left(f'| A(S)\right) \leq e^\varepsilon p\left(f'|A(S\backslash i)\right),
\end{equation}
for any $i \in S$ and where $A(\cdot)$ can take any value in some hypothesis class $\mathcal{H}$, given by e.g. the model architecture.
Since the exclusion of any data point cannot significantly change the value $f$ takes, the output of $f$ on any sample $i\in S$ can also not be significantly impacted.
This implies that changes in most metrics, due to the removal of a sample, (e.g. \cref{eq:self-inf} or \cref{eq:cross-infl}) will be small and thus memorisation will be bounded under DP. 

Concrete bounds on the memorisation ability (self-influence) of a differentially private model were previously presented in \cite{feldman2020does, van2021memorization}.
Further, using different interpretations of DP, it was also been shown that DP directly bounds the ability of an adversary to perform a MIA \cite{wasserman2010statistical} or training data reconstruction \cite{hayes2023bounding, stock2022defending}.
While applying DP to state-of-the-art ML models comes with a utility penalty (which may be unavoidable due to the fact that DP limits memorisation), recent works have demonstrated that, for many real-world datasets, competitive performance can be achieved by applying a number of training adaptations such as pre-training on large public datasets \cite{de2022unlocking, berrada2023unlocking}.
Overall, there is strong evidence that DP is the tool of choice for bounding the negative privacy implications and risks of memorisation in practice.

\begin{tcolorbox}[colback=gray!10!white, colframe=white!50!black, title=Key Point]
Differentially private ML training can (provably) prevent the negative traits caused by memorisation.  
\end{tcolorbox}

\section{Conclusion}
In this work, we systematically summarise and discuss the existing formulations of memorisation in machine learning models. 
We formulate the most widely used definitions and study their implications on the privacy of the data used to train the model. 
As quantifying memorisation using these definitions can be infeasible for many ML settings, we additionally discuss methods which allow the practitioners to estimate memorisation instead.
Furthermore, we discuss methods which allow the data owners to identify which specific samples are likely to be memorised using techniques from the domains of data valuation and influence estimation.
Finally, we outline the open challenges and questions to be collectively addressed by the machine learning community in order to be able to standardise the taxonomy in the field and permit more well-designed collaboration with privacy risks associated with model memorisation in mind. 
We hope that this work serves as a bridge between different ML communities and allows researchers to be more thoughtful when working with sensitive data.

\newpage

\bibliographystyle{ieeetr}
\bibliography{bibliography}
\vspace{12pt}

\end{document}

%% file: figs/long_tail.tex
\begin{tikzpicture}
    \begin{axis}[
        ylabel={Density},
        xmin=1, xmax=4,  
        ymin=0, ymax=0.75,  
        xtick=\empty,  
        width=10cm,
        height=6cm,
        xlabel style={
                yshift = {+4mm},
            },
        ylabel style={
                yshift = {-3mm}
            },
        legend pos=north east, 
        legend image code/.code={%
            \draw[#1, draw=none] (0cm,-0.1cm) rectangle (0.6cm,0.1cm);
            \draw[#1, ultra thick] (0cm,0cm) -- (0.6cm,0cm);
        }
    ]
    
    \addplot [
        domain=1:1.75,  
        samples=200,
        smooth,
        ultra thick,
        gray,
        fill=gray,  
        fill opacity=0.1  
    ]
    {x^(-3)} \closedcycle;
    \addlegendentry{Head}  

    \addplot [
        domain=1.75:5,  
        samples=200,
        smooth,
        ultra thick,  
        red,
        fill=red,  
        fill opacity=0.2  
    ]
    {x^(-3)} \closedcycle;
    \addlegendentry{Tail}  

    \addplot[dashed, black, very thick] coordinates {(1.75,0) (1.75,1.1)};

    \end{axis}
  \draw [gray, very thick, latex-latex] ([xshift=-1cm] 1.0, -0.4) -- ([xshift=-1cm]3.1, -0.4) node [midway, rotate=0, fill=white!92!gray, yshift=0pt] {typical};
  \draw [red, very thick, latex-latex] ([xshift=-1cm] 3.15, -0.4) -- ([xshift=-1cm]9.5, -0.4) node [midway, rotate=0, fill=white!95!red, yshift=0pt] {atypical} ;
    
\end{tikzpicture}

%% file: conference_101719.bbl
\begin{thebibliography}{100}

\bibitem{seo2020machine}
H.~Seo, M.~Badiei~Khuzani, V.~Vasudevan, C.~Huang, H.~Ren, R.~Xiao, X.~Jia, and L.~Xing, ``Machine learning techniques for biomedical image segmentation: an overview of technical aspects and introduction to state-of-art applications,'' {\em Medical physics}, vol.~47, no.~5, pp.~e148--e167, 2020.

\bibitem{rueckert2019model}
D.~Rueckert and J.~A. Schnabel, ``Model-based and data-driven strategies in medical image computing,'' {\em Proceedings of the IEEE}, vol.~108, no.~1, pp.~110--124, 2019.

\bibitem{nijkamp2022codegen}
E.~Nijkamp, B.~Pang, H.~Hayashi, L.~Tu, H.~Wang, Y.~Zhou, S.~Savarese, and C.~Xiong, ``Codegen: An open large language model for code with multi-turn program synthesis,'' {\em arXiv preprint arXiv:2203.13474}, 2022.

\bibitem{taylor2022galactica}
R.~Taylor, M.~Kardas, G.~Cucurull, T.~Scialom, A.~Hartshorn, E.~Saravia, A.~Poulton, V.~Kerkez, and R.~Stojnic, ``Galactica: A large language model for science,'' {\em arXiv preprint arXiv:2211.09085}, 2022.

\bibitem{alayrac2022flamingo}
J.-B. Alayrac, J.~Donahue, P.~Luc, A.~Miech, I.~Barr, Y.~Hasson, K.~Lenc, A.~Mensch, K.~Millican, M.~Reynolds, {\em et~al.}, ``Flamingo: a visual language model for few-shot learning,'' {\em Advances in Neural Information Processing Systems}, vol.~35, pp.~23716--23736, 2022.

\bibitem{ippolitopreventing}
D.~Ippolito, F.~Tram{\`e}r, M.~Nasr, C.~Zhang, M.~Jagielski, K.~Lee, C.~A. Choquette-Choo, and N.~Carlini, ``Preventing generation of verbatim memorization in language models gives a false sense of privacy,''

\bibitem{carlini2019secret}
N.~Carlini, C.~Liu, {\'U}.~Erlingsson, J.~Kos, and D.~Song, ``The secret sharer: Evaluating and testing unintended memorization in neural networks,'' in {\em 28th USENIX Security Symposium (USENIX Security 19)}, pp.~267--284, 2019.

\bibitem{carlini2021extracting}
N.~Carlini, F.~Tramer, E.~Wallace, M.~Jagielski, A.~Herbert-Voss, K.~Lee, A.~Roberts, T.~Brown, D.~Song, U.~Erlingsson, {\em et~al.}, ``Extracting training data from large language models,'' in {\em 30th USENIX Security Symposium (USENIX Security 21)}, pp.~2633--2650, 2021.

\bibitem{shokri2017membership}
R.~Shokri, M.~Stronati, C.~Song, and V.~Shmatikov, ``Membership inference attacks against machine learning models,'' in {\em 2017 IEEE symposium on security and privacy (SP)}, pp.~3--18, IEEE, 2017.

\bibitem{zhu2019deep}
L.~Zhu, Z.~Liu, and S.~Han, ``Deep leakage from gradients,'' {\em Advances in neural information processing systems}, vol.~32, 2019.

\bibitem{zhang2017understanding}
C.~Zhang, S.~Bengio, M.~Hardt, B.~Recht, and O.~Vinyals, ``Understanding deep learning requires rethinking generalization,'' 2017.

\bibitem{brown2021memorization}
G.~Brown, M.~Bun, V.~Feldman, A.~Smith, and K.~Talwar, ``When is memorization of irrelevant training data necessary for high-accuracy learning?,'' in {\em Proceedings of the 53rd annual ACM SIGACT symposium on theory of computing}, pp.~123--132, 2021.

\bibitem{feldman2020neural}
V.~Feldman and C.~Zhang, ``What neural networks memorize and why: Discovering the long tail via influence estimation,'' {\em Advances in Neural Information Processing Systems}, vol.~33, pp.~2881--2891, 2020.

\bibitem{feldman2020does}
V.~Feldman, ``Does learning require memorization? a short tale about a long tail,'' in {\em Proceedings of the 52nd Annual ACM SIGACT Symposium on Theory of Computing}, pp.~954--959, 2020.

\bibitem{zhang2021counterfactual}
C.~Zhang, D.~Ippolito, K.~Lee, M.~Jagielski, F.~Tram{\`e}r, and N.~Carlini, ``Counterfactual memorization in neural language models,'' {\em arXiv preprint arXiv:2112.12938}, 2021.

\bibitem{babbar2019data}
R.~Babbar and B.~Sch{\"o}lkopf, ``Data scarcity, robustness and extreme multi-label classification,'' {\em Machine Learning}, vol.~108, no.~8-9, pp.~1329--1351, 2019.

\bibitem{zhu2014capturing}
X.~Zhu, D.~Anguelov, and D.~Ramanan, ``Capturing long-tail distributions of object subcategories,'' in {\em Proceedings of the IEEE Conference on Computer Vision and Pattern Recognition}, pp.~915--922, 2014.

\bibitem{van2017devil}
G.~Van~Horn and P.~Perona, ``The devil is in the tails: Fine-grained classification in the wild,'' {\em arXiv preprint arXiv:1709.01450}, 2017.

\bibitem{yang2022survey}
L.~Yang, H.~Jiang, Q.~Song, and J.~Guo, ``A survey on long-tailed visual recognition,'' {\em International Journal of Computer Vision}, vol.~130, no.~7, pp.~1837--1872, 2022.

\bibitem{olatunji2021membership}
I.~E. Olatunji, W.~Nejdl, and M.~Khosla, ``Membership inference attack on graph neural networks,'' in {\em 2021 Third IEEE International Conference on Trust, Privacy and Security in Intelligent Systems and Applications (TPS-ISA)}, pp.~11--20, IEEE, 2021.

\bibitem{chen2020gan}
D.~Chen, N.~Yu, Y.~Zhang, and M.~Fritz, ``Gan-leaks: A taxonomy of membership inference attacks against generative models,'' in {\em Proceedings of the 2020 ACM SIGSAC conference on computer and communications security}, pp.~343--362, 2020.

\bibitem{kuppa2021towards}
A.~Kuppa, L.~Aouad, and N.-A. Le-Khac, ``Towards improving privacy of synthetic datasets,'' in {\em Annual Privacy Forum}, pp.~106--119, Springer, 2021.

\bibitem{leino2020stolen}
K.~Leino and M.~Fredrikson, ``Stolen memories: Leveraging model memorization for calibrated $\{$White-Box$\}$ membership inference,'' in {\em 29th USENIX security symposium (USENIX Security 20)}, pp.~1605--1622, 2020.

\bibitem{veale2018algorithms}
M.~Veale, R.~Binns, and L.~Edwards, ``Algorithms that remember: model inversion attacks and data protection law,'' {\em Philosophical Transactions of the Royal Society A: Mathematical, Physical and Engineering Sciences}, vol.~376, no.~2133, p.~20180083, 2018.

\bibitem{hilprecht2019monte}
B.~Hilprecht, M.~H{\"a}rterich, and D.~Bernau, ``Monte carlo and reconstruction membership inference attacks against generative models.,'' {\em Proc. Priv. Enhancing Technol.}, vol.~2019, no.~4, pp.~232--249, 2019.

\bibitem{mehta2020extreme}
H.~Mehta, A.~Cutkosky, and B.~Neyshabur, ``Extreme memorization via scale of initialization,'' {\em arXiv preprint arXiv:2008.13363}, 2020.

\bibitem{tirumala2022memorization}
K.~Tirumala, A.~Markosyan, L.~Zettlemoyer, and A.~Aghajanyan, ``Memorization without overfitting: Analyzing the training dynamics of large language models,'' {\em Advances in Neural Information Processing Systems}, vol.~35, pp.~38274--38290, 2022.

\bibitem{cook1977detection}
R.~D. Cook, ``Detection of influential observation in linear regression,'' {\em Technometrics}, vol.~19, no.~1, pp.~15--18, 1977.

\bibitem{hammoudeh2022training}
Z.~Hammoudeh and D.~Lowd, ``Training data influence analysis and estimation: A survey,'' {\em arXiv preprint arXiv:2212.04612}, 2022.

\bibitem{koh2017understanding}
P.~W. Koh and P.~Liang, ``Understanding black-box predictions via influence functions,'' in {\em International conference on machine learning}, pp.~1885--1894, PMLR, 2017.

\bibitem{cook1982residuals}
R.~D. Cook and S.~Weisberg, ``Residuals and influence in regression,'' 1982.

\bibitem{guo2020fastif}
H.~Guo, N.~F. Rajani, P.~Hase, M.~Bansal, and C.~Xiong, ``Fastif: Scalable influence functions for efficient model interpretation and debugging,'' {\em arXiv preprint arXiv:2012.15781}, 2020.

\bibitem{schioppa2022scaling}
A.~Schioppa, P.~Zablotskaia, D.~Vilar, and A.~Sokolov, ``Scaling up influence functions,'' in {\em Proceedings of the AAAI Conference on Artificial Intelligence}, vol.~36, pp.~8179--8186, 2022.

\bibitem{grosse2023studying}
R.~Grosse, J.~Bae, C.~Anil, N.~Elhage, A.~Tamkin, A.~Tajdini, B.~Steiner, D.~Li, E.~Durmus, E.~Perez, {\em et~al.}, ``Studying large language model generalization with influence functions,'' {\em arXiv preprint arXiv:2308.03296}, 2023.

\bibitem{bae2022if}
J.~Bae, N.~Ng, A.~Lo, M.~Ghassemi, and R.~B. Grosse, ``If influence functions are the answer, then what is the question?,'' {\em Advances in Neural Information Processing Systems}, vol.~35, pp.~17953--17967, 2022.

\bibitem{basu2020influence}
S.~Basu, P.~Pope, and S.~Feizi, ``Influence functions in deep learning are fragile,'' in {\em International Conference on Learning Representations}, 2020.

\bibitem{schioppa2023theoretical}
A.~Schioppa, K.~Filippova, I.~Titov, and P.~Zablotskaia, ``Theoretical and practical perspectives on what influence functions do,'' {\em arXiv preprint arXiv:2305.16971}, 2023.

\bibitem{yeh2018representer}
C.-K. Yeh, J.~Kim, I.~E.-H. Yen, and P.~K. Ravikumar, ``Representer point selection for explaining deep neural networks,'' {\em Advances in neural information processing systems}, vol.~31, 2018.

\bibitem{yeh2022first}
C.-K. Yeh, A.~Taly, M.~Sundararajan, F.~Liu, and P.~Ravikumar, ``First is better than last for language data influence,'' {\em Advances in Neural Information Processing Systems}, vol.~35, pp.~32285--32298, 2022.

\bibitem{hammoudeh2022identifying}
Z.~Hammoudeh and D.~Lowd, ``Identifying a training-set attack's target using renormalized influence estimation,'' in {\em Proceedings of the 2022 ACM SIGSAC Conference on Computer and Communications Security}, pp.~1367--1381, 2022.

\bibitem{ghorbani2019data}
A.~Ghorbani and J.~Zou, ``Data shapley: Equitable valuation of data for machine learning,'' in {\em International conference on machine learning}, pp.~2242--2251, PMLR, 2019.

\bibitem{chen2020optimal}
W.~Chen, S.~Horvath, and P.~Richtarik, ``Optimal client sampling for federated learning,'' {\em arXiv preprint arXiv:2010.13723}, 2020.

\bibitem{amiri2021convergence}
M.~M. Amiri, D.~G{\"u}nd{\"u}z, S.~R. Kulkarni, and H.~V. Poor, ``Convergence of update aware device scheduling for federated learning at the wireless edge,'' {\em IEEE Transactions on Wireless Communications}, vol.~20, no.~6, pp.~3643--3658, 2021.

\bibitem{lai2021oort}
F.~Lai, X.~Zhu, H.~V. Madhyastha, and M.~Chowdhury, ``Oort: Efficient federated learning via guided participant selection,'' in {\em 15th $\{$USENIX$\}$ Symposium on Operating Systems Design and Implementation ($\{$OSDI$\}$ 21)}, pp.~19--35, 2021.

\bibitem{zhu2022isfl}
Z.~Zhu, P.~Fan, C.~Peng, and K.~B. Letaief, ``Isfl: Trustworthy federated learning for non-iid data with local importance sampling,'' {\em arXiv preprint arXiv:2210.02119}, 2022.

\bibitem{katharopoulos2018not}
A.~Katharopoulos and F.~Fleuret, ``Not all samples are created equal: Deep learning with importance sampling,'' in {\em International conference on machine learning}, pp.~2525--2534, PMLR, 2018.

\bibitem{li2021sample}
A.~Li, L.~Zhang, J.~Tan, Y.~Qin, J.~Wang, and X.-Y. Li, ``Sample-level data selection for federated learning,'' in {\em IEEE INFOCOM 2021-IEEE Conference on Computer Communications}, pp.~1--10, IEEE, 2021.

\bibitem{xue2021toward}
Y.~Xue, C.~Niu, Z.~Zheng, S.~Tang, C.~Lyu, F.~Wu, and G.~Chen, ``Toward understanding the influence of individual clients in federated learning,'' in {\em Proceedings of the AAAI Conference on Artificial Intelligence}, vol.~35, pp.~10560--10567, 2021.

\bibitem{Abadi2016}
M.~Abadi, A.~Chu, I.~Goodfellow, H.~B. McMahan, I.~Mironov, K.~Talwar, and L.~Zhang, ``Deep learning with differential privacy,'' in {\em Proceedings of the 2016 {ACM} {SIGSAC} Conference on Computer and Communications Security}, {ACM}, Oct. 2016.

\bibitem{feldman2021individual}
V.~Feldman and T.~Zrnic, ``Individual privacy accounting via a renyi filter,'' {\em Advances in Neural Information Processing Systems}, vol.~34, pp.~28080--28091, 2021.

\bibitem{yu2023individual}
D.~Yu, G.~Kamath, J.~Kulkarni, T.-Y. Liu, J.~Yin, and H.~Zhang, ``Individual privacy accounting for differentially private stochastic gradient descent,'' {\em Transactions on Machine Learning Research}, 2023.

\bibitem{koskela2022individual}
A.~Koskela, M.~Tobaben, and A.~Honkela, ``Individual privacy accounting with gaussian differential privacy,'' {\em arXiv preprint arXiv:2209.15596}, 2022.

\bibitem{usynin2023beyond}
D.~Usynin, D.~Rueckert, and G.~Kaissis, ``Beyond gradients: Exploiting adversarial priors in model inversion attacks,'' {\em ACM Transactions on Privacy and Security}, vol.~26, no.~3, pp.~1--30, 2023.

\bibitem{geiping2020inverting}
J.~Geiping, H.~Bauermeister, H.~Dr{\"o}ge, and M.~Moeller, ``Inverting gradients-how easy is it to break privacy in federated learning?,'' {\em Advances in Neural Information Processing Systems}, vol.~33, pp.~16937--16947, 2020.

\bibitem{balle2022reconstructing}
B.~Balle, G.~Cherubin, and J.~Hayes, ``Reconstructing training data with informed adversaries,'' in {\em 2022 IEEE Symposium on Security and Privacy (SP)}, pp.~1138--1156, IEEE, 2022.

\bibitem{garg2023memorization}
I.~Garg and K.~Roy, ``Memorization through the lens of curvature of loss function around samples,'' {\em arXiv preprint arXiv:2307.05831}, 2023.

\bibitem{pruthi2020estimating}
G.~Pruthi, F.~Liu, S.~Kale, and M.~Sundararajan, ``Estimating training data influence by tracing gradient descent,'' {\em Advances in Neural Information Processing Systems}, vol.~33, pp.~19920--19930, 2020.

\bibitem{agarwal2022estimating}
C.~Agarwal, D.~D'souza, and S.~Hooker, ``Estimating example difficulty using variance of gradients,'' in {\em Proceedings of the IEEE/CVF Conference on Computer Vision and Pattern Recognition}, pp.~10368--10378, 2022.

\bibitem{mueller2022input}
T.~T. Mueller, S.~Kolek, F.~Jungmann, A.~Ziller, D.~Usynin, M.~Knolle, D.~Rueckert, and G.~Kaissis, ``How do input attributes impact the privacy loss in differential privacy?,'' {\em arXiv preprint arXiv:2211.10173}, 2022.

\bibitem{hannun2021measuring}
A.~Hannun, C.~Guo, and L.~van~der Maaten, ``Measuring data leakage in machine-learning models with fisher information,'' in {\em Uncertainty in Artificial Intelligence}, pp.~760--770, PMLR, 2021.

\bibitem{kolmogorov1956shannon}
A.~Kolmogorov, ``On the shannon theory of information transmission in the case of continuous signals,'' {\em IRE Transactions on Information Theory}, vol.~2, no.~4, pp.~102--108, 1956.

\bibitem{shwartz2022information}
R.~Shwartz-Ziv, ``Information flow in deep neural networks,'' {\em arXiv preprint arXiv:2202.06749}, 2022.

\bibitem{goldfeld2018estimating}
Z.~Goldfeld, E.~v.~d. Berg, K.~Greenewald, I.~Melnyk, N.~Nguyen, B.~Kingsbury, and Y.~Polyanskiy, ``Estimating information flow in deep neural networks,'' {\em arXiv preprint arXiv:1810.05728}, 2018.

\bibitem{xu2020theory}
Y.~Xu, S.~Zhao, J.~Song, R.~Stewart, and S.~Ermon, ``A theory of usable information under computational constraints,'' {\em arXiv preprint arXiv:2002.10689}, 2020.

\bibitem{ethayarajh2022understanding}
K.~Ethayarajh, Y.~Choi, and S.~Swayamdipta, ``Understanding dataset difficulty with $v$-usable information,'' in {\em International Conference on Machine Learning}, pp.~5988--6008, PMLR, 2022.

\bibitem{mo2021quantifying}
F.~Mo, A.~Borovykh, M.~Malekzadeh, H.~Haddadi, and S.~Demetriou, ``Quantifying information leakage from gradients,'' {\em CoRR, abs/2105.13929}, 2021.

\bibitem{zhao2019inferring}
B.~Z.~H. Zhao, H.~J. Asghar, R.~Bhaskar, and M.~A. Kaafar, ``On inferring training data attributes in machine learning models,'' {\em arXiv preprint arXiv:1908.10558}, 2019.

\bibitem{goldfeld2021sliced}
Z.~Goldfeld and K.~Greenewald, ``Sliced mutual information: A scalable measure of statistical dependence,'' {\em Advances in Neural Information Processing Systems}, vol.~34, pp.~17567--17578, 2021.

\bibitem{wongso2023using}
S.~Wongso, R.~Ghosh, and M.~Motani, ``Using sliced mutual information to study memorization and generalization in deep neural networks,'' in {\em International Conference on Artificial Intelligence and Statistics}, pp.~11608--11629, PMLR, 2023.

\bibitem{harutyunyan2021estimating}
H.~Harutyunyan, A.~Achille, G.~Paolini, O.~Majumder, A.~Ravichandran, R.~Bhotika, and S.~Soatto, ``Estimating informativeness of samples with smooth unique information,'' {\em arXiv preprint arXiv:2101.06640}, 2021.

\bibitem{usynin2023leveraging}
D.~Usynin, D.~Rueckert, and G.~Kaissis, ``Leveraging gradient-derived metrics for data selection and valuation in differentially private training,'' {\em arXiv preprint arXiv:2305.02942}, 2023.

\bibitem{chen2021synthetic}
R.~J. Chen, M.~Y. Lu, T.~Y. Chen, D.~F. Williamson, and F.~Mahmood, ``Synthetic data in machine learning for medicine and healthcare,'' {\em Nature Biomedical Engineering}, vol.~5, no.~6, pp.~493--497, 2021.

\bibitem{baldock2021deep}
R.~Baldock, H.~Maennel, and B.~Neyshabur, ``Deep learning through the lens of example difficulty,'' {\em Advances in Neural Information Processing Systems}, vol.~34, pp.~10876--10889, 2021.

\bibitem{garg2023samples}
I.~Garg and K.~Roy, ``Samples with low loss curvature improve data efficiency,'' in {\em Proceedings of the IEEE/CVF Conference on Computer Vision and Pattern Recognition}, pp.~20290--20300, 2023.

\bibitem{carlini2019distribution}
N.~Carlini, U.~Erlingsson, and N.~Papernot, ``Distribution density, tails, and outliers in machine learning: Metrics and applications,'' {\em arXiv preprint arXiv:1910.13427}, 2019.

\bibitem{bau2017network}
D.~Bau, B.~Zhou, A.~Khosla, A.~Oliva, and A.~Torralba, ``Network dissection: Quantifying interpretability of deep visual representations,'' in {\em Proceedings of the IEEE conference on computer vision and pattern recognition}, pp.~6541--6549, 2017.

\bibitem{maini2023can}
P.~Maini, M.~C. Mozer, H.~Sedghi, Z.~C. Lipton, J.~Z. Kolter, and C.~Zhang, ``Can neural network memorization be localized?,'' {\em arXiv preprint arXiv:2307.09542}, 2023.

\bibitem{stephenson2021geometry}
C.~Stephenson, S.~Padhy, A.~Ganesh, Y.~Hui, H.~Tang, and S.~Chung, ``On the geometry of generalization and memorization in deep neural networks,'' {\em arXiv preprint arXiv:2105.14602}, 2021.

\bibitem{arpit2017closer}
D.~Arpit, S.~Jastrzkebski, N.~Ballas, D.~Krueger, E.~Bengio, M.~S. Kanwal, T.~Maharaj, A.~Fischer, A.~Courville, Y.~Bengio, {\em et~al.}, ``A closer look at memorization in deep networks,'' in {\em International conference on machine learning}, pp.~233--242, PMLR, 2017.

\bibitem{kishida2019empirical}
I.~Kishida and H.~Nakayama, ``Empirical study of easy and hard examples in cnn training,'' in {\em Neural Information Processing: 26th International Conference, ICONIP 2019, Sydney, NSW, Australia, December 12--15, 2019, Proceedings, Part IV 26}, pp.~179--188, Springer, 2019.

\bibitem{paul2021deep}
M.~Paul, S.~Ganguli, and G.~K. Dziugaite, ``Deep learning on a data diet: Finding important examples early in training,'' {\em Advances in Neural Information Processing Systems}, vol.~34, pp.~20596--20607, 2021.

\bibitem{thakkar2020understanding}
O.~Thakkar, S.~Ramaswamy, R.~Mathews, and F.~Beaufays, ``Understanding unintended memorization in federated learning,'' {\em arXiv preprint arXiv:2006.07490}, 2020.

\bibitem{lee2021deduplicating}
K.~Lee, D.~Ippolito, A.~Nystrom, C.~Zhang, D.~Eck, C.~Callison-Burch, and N.~Carlini, ``Deduplicating training data makes language models better,'' {\em arXiv preprint arXiv:2107.06499}, 2021.

\bibitem{hartley2022measuring}
J.~Hartley and S.~A. Tsaftaris, ``Measuring unintended memorisation of unique private features in neural networks,'' {\em arXiv preprint arXiv:2202.08099}, 2022.

\bibitem{liang2023code}
J.~Liang, W.~Huang, F.~Xia, P.~Xu, K.~Hausman, B.~Ichter, P.~Florence, and A.~Zeng, ``Code as policies: Language model programs for embodied control,'' in {\em 2023 IEEE International Conference on Robotics and Automation (ICRA)}, pp.~9493--9500, IEEE, 2023.

\bibitem{wei2022emergent}
J.~Wei, Y.~Tay, R.~Bommasani, C.~Raffel, B.~Zoph, S.~Borgeaud, D.~Yogatama, M.~Bosma, D.~Zhou, D.~Metzler, {\em et~al.}, ``Emergent abilities of large language models,'' {\em arXiv preprint arXiv:2206.07682}, 2022.

\bibitem{carlini2022quantifying}
N.~Carlini, D.~Ippolito, M.~Jagielski, K.~Lee, F.~Tramer, and C.~Zhang, ``Quantifying memorization across neural language models,'' {\em arXiv preprint arXiv:2202.07646}, 2022.

\bibitem{carlini2023extracting}
N.~Carlini, J.~Hayes, M.~Nasr, M.~Jagielski, V.~Sehwag, F.~Tramer, B.~Balle, D.~Ippolito, and E.~Wallace, ``Extracting training data from diffusion models,'' in {\em 32nd USENIX Security Symposium (USENIX Security 23)}, pp.~5253--5270, 2023.

\bibitem{chobola2022membership}
T.~Chobola, D.~Usynin, and G.~Kaissis, ``Membership inference attacks against semantic segmentation models,'' {\em arXiv preprint arXiv:2212.01082}, 2022.

\bibitem{zhou2021deep}
X.~Zhou, M.~Xu, Y.~Wu, and N.~Zheng, ``Deep model poisoning attack on federated learning,'' {\em Future Internet}, vol.~13, no.~3, p.~73, 2021.

\bibitem{usynin2021adversarial}
D.~Usynin, A.~Ziller, M.~Makowski, R.~Braren, D.~Rueckert, B.~Glocker, G.~Kaissis, and J.~Passerat-Palmbach, ``Adversarial interference and its mitigations in privacy-preserving collaborative machine learning,'' {\em Nature Machine Intelligence}, vol.~3, no.~9, pp.~749--758, 2021.

\bibitem{tramer2022truth}
F.~Tram{\`e}r, R.~Shokri, A.~San~Joaquin, H.~Le, M.~Jagielski, S.~Hong, and N.~Carlini, ``Truth serum: Poisoning machine learning models to reveal their secrets,'' in {\em Proceedings of the 2022 ACM SIGSAC Conference on Computer and Communications Security}, pp.~2779--2792, 2022.

\bibitem{carlini2022privacy}
N.~Carlini, M.~Jagielski, C.~Zhang, N.~Papernot, A.~Terzis, and F.~Tramer, ``The privacy onion effect: Memorization is relative,'' {\em Advances in Neural Information Processing Systems}, vol.~35, pp.~13263--13276, 2022.

\bibitem{bagdasaryan2021blind}
E.~Bagdasaryan and V.~Shmatikov, ``Blind backdoors in deep learning models,'' in {\em 30th USENIX Security Symposium (USENIX Security 21)}, pp.~1505--1521, 2021.

\bibitem{tian2022comprehensive}
Z.~Tian, L.~Cui, J.~Liang, and S.~Yu, ``A comprehensive survey on poisoning attacks and countermeasures in machine learning,'' {\em ACM Computing Surveys}, vol.~55, no.~8, pp.~1--35, 2022.

\bibitem{sablayrolles2019white}
A.~Sablayrolles, M.~Douze, C.~Schmid, Y.~Ollivier, and H.~J{\'e}gou, ``White-box vs black-box: Bayes optimal strategies for membership inference,'' in {\em International Conference on Machine Learning}, pp.~5558--5567, PMLR, 2019.

\bibitem{choquette2021label}
C.~A. Choquette-Choo, F.~Tramer, N.~Carlini, and N.~Papernot, ``Label-only membership inference attacks,'' in {\em International conference on machine learning}, pp.~1964--1974, PMLR, 2021.

\bibitem{liu2019performing}
K.~S. Liu, C.~Xiao, B.~Li, and J.~Gao, ``Performing co-membership attacks against deep generative models,'' in {\em 2019 IEEE International Conference on Data Mining (ICDM)}, pp.~459--467, IEEE, 2019.

\bibitem{choi2023train}
J.~Choi, S.~Tople, V.~Chandrasekaran, and S.~Jha, ``Why train more? effective and efficient membership inference via memorization,'' {\em arXiv preprint arXiv:2310.08015}, 2023.

\bibitem{usynin2022zen}
D.~Usynin, D.~Rueckert, J.~Passerat-Palmbach, and G.~Kaissis, ``Zen and the art of model adaptation: Low-utility-cost attack mitigations in collaborative machine learning.,'' {\em Proc. Priv. Enhancing Technol.}, vol.~2022, no.~1, pp.~274--290, 2022.

\bibitem{truex2019demystifying}
S.~Truex, L.~Liu, M.~E. Gursoy, L.~Yu, and W.~Wei, ``Demystifying membership inference attacks in machine learning as a service,'' {\em IEEE Transactions on Services Computing}, vol.~14, no.~6, pp.~2073--2089, 2019.

\bibitem{salem2018ml}
A.~Salem, Y.~Zhang, M.~Humbert, P.~Berrang, M.~Fritz, and M.~Backes, ``Ml-leaks: Model and data independent membership inference attacks and defenses on machine learning models,'' {\em arXiv preprint arXiv:1806.01246}, 2018.

\bibitem{carlini2022membership}
N.~Carlini, S.~Chien, M.~Nasr, S.~Song, A.~Terzis, and F.~Tramer, ``Membership inference attacks from first principles,'' in {\em 2022 IEEE Symposium on Security and Privacy (SP)}, pp.~1897--1914, IEEE, 2022.

\bibitem{yeom2018privacy}
S.~Yeom, I.~Giacomelli, M.~Fredrikson, and S.~Jha, ``Privacy risk in machine learning: Analyzing the connection to overfitting,'' in {\em 2018 IEEE 31st computer security foundations symposium (CSF)}, pp.~268--282, IEEE, 2018.

\bibitem{jayaraman2020revisiting}
B.~Jayaraman, L.~Wang, K.~Knipmeyer, Q.~Gu, and D.~Evans, ``Revisiting membership inference under realistic assumptions,'' {\em arXiv preprint arXiv:2005.10881}, 2020.

\bibitem{irolla2019demystifying}
P.~Irolla and G.~Ch{\^a}tel, ``Demystifying the membership inference attack,'' in {\em 2019 12th CMI Conference on Cybersecurity and Privacy (CMI)}, pp.~1--7, IEEE, 2019.

\bibitem{bentley2020quantifying}
J.~W. Bentley, D.~Gibney, G.~Hoppenworth, and S.~K. Jha, ``Quantifying membership inference vulnerability via generalization gap and other model metrics,'' {\em arXiv preprint arXiv:2009.05669}, 2020.

\bibitem{li2022privacy}
X.~Li, Q.~Li, Z.~Hu, and X.~Hu, ``On the privacy effect of data enhancement via the lens of memorization,'' {\em arXiv preprint arXiv:2208.08270}, 2022.

\bibitem{watson2021importance}
L.~Watson, C.~Guo, G.~Cormode, and A.~Sablayrolles, ``On the importance of difficulty calibration in membership inference attacks,'' {\em arXiv preprint arXiv:2111.08440}, 2021.

\bibitem{gu2022cs}
Y.~Gu, Y.~Bai, and S.~Xu, ``Cs-mia: Membership inference attack based on prediction confidence series in federated learning,'' {\em Journal of Information Security and Applications}, vol.~67, p.~103201, 2022.

\bibitem{hayes2017logan}
J.~Hayes, L.~Melis, G.~Danezis, and E.~De~Cristofaro, ``Logan: Membership inference attacks against generative models,'' {\em arXiv preprint arXiv:1705.07663}, 2017.

\bibitem{jayaraman2019evaluating}
B.~Jayaraman and D.~Evans, ``Evaluating differentially private machine learning in practice,'' in {\em 28th USENIX Security Symposium (USENIX Security 19)}, pp.~1895--1912, 2019.

\bibitem{song2021systematic}
L.~Song and P.~Mittal, ``Systematic evaluation of privacy risks of machine learning models,'' in {\em 30th USENIX Security Symposium (USENIX Security 21)}, pp.~2615--2632, 2021.

\bibitem{kosinski2013private}
M.~Kosinski, D.~Stillwell, and T.~Graepel, ``Private traits and attributes are predictable from digital records of human behavior,'' {\em Proceedings of the national academy of sciences}, vol.~110, no.~15, pp.~5802--5805, 2013.

\bibitem{feng2021attribute}
T.~Feng, H.~Hashemi, R.~Hebbar, M.~Annavaram, and S.~S. Narayanan, ``Attribute inference attack of speech emotion recognition in federated learning settings,'' {\em arXiv preprint arXiv:2112.13416}, 2021.

\bibitem{jourdan2021privacy}
T.~Jourdan, A.~Boutet, and C.~Frindel, ``Privacy assessment of federated learning using private personalized layers,'' in {\em 2021 IEEE 31st International Workshop on Machine Learning for Signal Processing (MLSP)}, pp.~1--6, IEEE, 2021.

\bibitem{thomas2020investigating}
A.~Thomas, D.~I. Adelani, A.~Davody, A.~Mogadala, and D.~Klakow, ``Investigating the impact of pre-trained word embeddings on memorization in neural networks,'' in {\em Text, Speech, and Dialogue: 23rd International Conference, TSD 2020, Brno, Czech Republic, September 8--11, 2020, Proceedings 23}, pp.~273--281, Springer, 2020.

\bibitem{haim2022reconstructing}
N.~Haim, G.~Vardi, G.~Yehudai, O.~Shamir, and M.~Irani, ``Reconstructing training data from trained neural networks,'' {\em Advances in Neural Information Processing Systems}, vol.~35, pp.~22911--22924, 2022.

\bibitem{fredrikson2015model}
M.~Fredrikson, S.~Jha, and T.~Ristenpart, ``Model inversion attacks that exploit confidence information and basic countermeasures,'' in {\em Proceedings of the 22nd ACM SIGSAC conference on computer and communications security}, pp.~1322--1333, 2015.

\bibitem{ye2022enhanced}
J.~Ye, A.~Maddi, S.~K. Murakonda, V.~Bindschaedler, and R.~Shokri, ``Enhanced membership inference attacks against machine learning models,'' in {\em Proceedings of the 2022 ACM SIGSAC Conference on Computer and Communications Security}, pp.~3093--3106, 2022.

\bibitem{nasr2021adversary}
M.~Nasr, S.~Songi, A.~Thakurta, N.~Papernot, and N.~Carlin, ``Adversary instantiation: Lower bounds for differentially private machine learning,'' in {\em 2021 IEEE Symposium on security and privacy (SP)}, pp.~866--882, IEEE, 2021.

\bibitem{kulynych2019disparate}
B.~Kulynych, M.~Yaghini, G.~Cherubin, M.~Veale, and C.~Troncoso, ``Disparate vulnerability to membership inference attacks,'' {\em arXiv preprint arXiv:1906.00389}, 2019.

\bibitem{liu2019ppgan}
Y.~Liu, J.~Peng, J.~James, and Y.~Wu, ``Ppgan: Privacy-preserving generative adversarial network,'' in {\em 2019 IEEE 25Th international conference on parallel and distributed systems (ICPADS)}, pp.~985--989, IEEE, 2019.

\bibitem{triastcyn2020federated}
A.~Triastcyn and B.~Faltings, ``Federated generative privacy,'' {\em IEEE Intelligent Systems}, vol.~35, no.~4, pp.~50--57, 2020.

\bibitem{wu2019generalization}
B.~Wu, S.~Zhao, C.~Chen, H.~Xu, L.~Wang, X.~Zhang, G.~Sun, and J.~Zhou, ``Generalization in generative adversarial networks: A novel perspective from privacy protection,'' {\em Advances in Neural Information Processing Systems}, vol.~32, 2019.

\bibitem{fernandez2023privacy}
V.~Fernandez, P.~Sanchez, W.~H.~L. Pinaya, G.~Jacenk{\'o}w, S.~A. Tsaftaris, and J.~Cardoso, ``Privacy distillation: Reducing re-identification risk of multimodal diffusion models,'' {\em arXiv preprint arXiv:2306.01322}, 2023.

\bibitem{packhauser2022deep}
K.~Packh{\"a}user, S.~G{\"u}ndel, N.~M{\"u}nster, C.~Syben, V.~Christlein, and A.~Maier, ``Deep learning-based patient re-identification is able to exploit the biometric nature of medical chest x-ray data,'' {\em Scientific Reports}, vol.~12, no.~1, p.~14851, 2022.

\bibitem{toneva2018empirical}
M.~Toneva, A.~Sordoni, R.~T.~d. Combes, A.~Trischler, Y.~Bengio, and G.~J. Gordon, ``An empirical study of example forgetting during deep neural network learning,'' {\em arXiv preprint arXiv:1812.05159}, 2018.

\bibitem{jagielski2022measuring}
M.~Jagielski, O.~Thakkar, F.~Tramer, D.~Ippolito, K.~Lee, N.~Carlini, E.~Wallace, S.~Song, A.~Thakurta, N.~Papernot, {\em et~al.}, ``Measuring forgetting of memorized training examples,'' {\em arXiv preprint arXiv:2207.00099}, 2022.

\bibitem{bourtoule2021machine}
L.~Bourtoule, V.~Chandrasekaran, C.~A. Choquette-Choo, H.~Jia, A.~Travers, B.~Zhang, D.~Lie, and N.~Papernot, ``Machine unlearning,'' in {\em 2021 IEEE Symposium on Security and Privacy (SP)}, pp.~141--159, IEEE, 2021.

\bibitem{zhou2022fortuitous}
H.~Zhou, A.~Vani, H.~Larochelle, and A.~Courville, ``Fortuitous forgetting in connectionist networks,'' {\em arXiv preprint arXiv:2202.00155}, 2022.

\bibitem{dpbook}
C.~Dwork and A.~Roth, ``The algorithmic foundations of differential privacy,'' {\em Foundations and Trends® in Theoretical Computer Science}, vol.~9, no.~3–4, pp.~211--407, 2014.

\bibitem{van2021memorization}
G.~van~den Burg and C.~Williams, ``On memorization in probabilistic deep generative models,'' {\em Advances in Neural Information Processing Systems}, vol.~34, pp.~27916--27928, 2021.

\bibitem{wasserman2010statistical}
L.~Wasserman and S.~Zhou, ``A statistical framework for differential privacy,'' {\em Journal of the American Statistical Association}, vol.~105, no.~489, pp.~375--389, 2010.

\bibitem{hayes2023bounding}
J.~Hayes, S.~Mahloujifar, and B.~Balle, ``Bounding training data reconstruction in dp-sgd,'' {\em arXiv preprint arXiv:2302.07225}, 2023.

\bibitem{stock2022defending}
P.~Stock, I.~Shilov, I.~Mironov, and A.~Sablayrolles, ``Defending against reconstruction attacks with r$\backslash$'enyi differential privacy,'' {\em arXiv preprint arXiv:2202.07623}, 2022.

\bibitem{de2022unlocking}
S.~De, L.~Berrada, J.~Hayes, S.~L. Smith, and B.~Balle, ``Unlocking high-accuracy differentially private image classification through scale,'' {\em arXiv preprint arXiv:2204.13650}, 2022.

\bibitem{berrada2023unlocking}
L.~Berrada, S.~De, J.~H. Shen, J.~Hayes, R.~Stanforth, D.~Stutz, P.~Kohli, S.~L. Smith, and B.~Balle, ``Unlocking accuracy and fairness in differentially private image classification,'' {\em arXiv preprint arXiv:2308.10888}, 2023.

\end{thebibliography}
